\documentclass{IEEEtran}

\usepackage[utf8]{inputenc} 
\usepackage[T1]{fontenc}    
\usepackage{hyperref}       
\usepackage{url}            
\usepackage{booktabs}       
\usepackage{amsfonts}       
\usepackage{nicefrac}       
\usepackage{microtype}      
\usepackage{subfiles}
\usepackage{graphicx, subcaption}
\usepackage{amsmath}
\usepackage{wrapfig}
\usepackage{multirow}
\usepackage{floatrow}
\usepackage{booktabs}
\usepackage{subcaption}
\DeclareMathOperator*{\argmax}{arg\,max}

\title{Semantic Example Guided Image-to-Image Translation}

\author{%
    Jialu Huang, Jing Liao, Tak Wu Sam Kwong\\
  Department of Computer Science\\
  City University of Hong Kong\\
}

\begin{document}
\maketitle

\begin{abstract}
Many image-to-image (I2I) translation problems are in nature of high diversity that a single input may have various counterparts. The multi-modal network that can build many-to-many mapping between two visual domains has been proposed in prior works. However, most of them are guided by sampled noises. Some others encode the reference images into a latent vector, by which the semantic information of the reference image are washed away. In this work, we aim to provide a solution to control the output based on references semantically. Given a reference image and an input in another domain, we first perform semantic matching between the two visual content and generate the auxiliary image, which explicitly encourages the semantic characteristic to be preserved. A deep network then is used for I2I translation and the final outputs are expected to be semantically similar to both the input and the reference. However, few paired data can satisfy that dual-similarity in a supervised fashion, and so we build up a self-supervised framework to serve the training purpose. We improve the quality and diversity of the outputs by employing non-local blocks and multi-task architecture. We assess the proposed method through extensive qualitative and quantitative evaluations and also present comparisons with several state-of-the-art models.
\end{abstract}

\section{Introduction}
Image-to-image(I2I) translation by deep neural networks will shed light on many vision and graphic applications, such as image synthesis (sketch/label/artwork to photos), colorization (grayscale to the color), image enhancement (low-resolution to high resolution), etc. Deep neural networks that can learn mapping between two visual domains and generate images in the target domain with a certain level of diversity could become a set of powerful tools in industrial design, digital art and animation/game production.\par

\begin{figure}
    \centering
    \includegraphics[width=\textwidth]{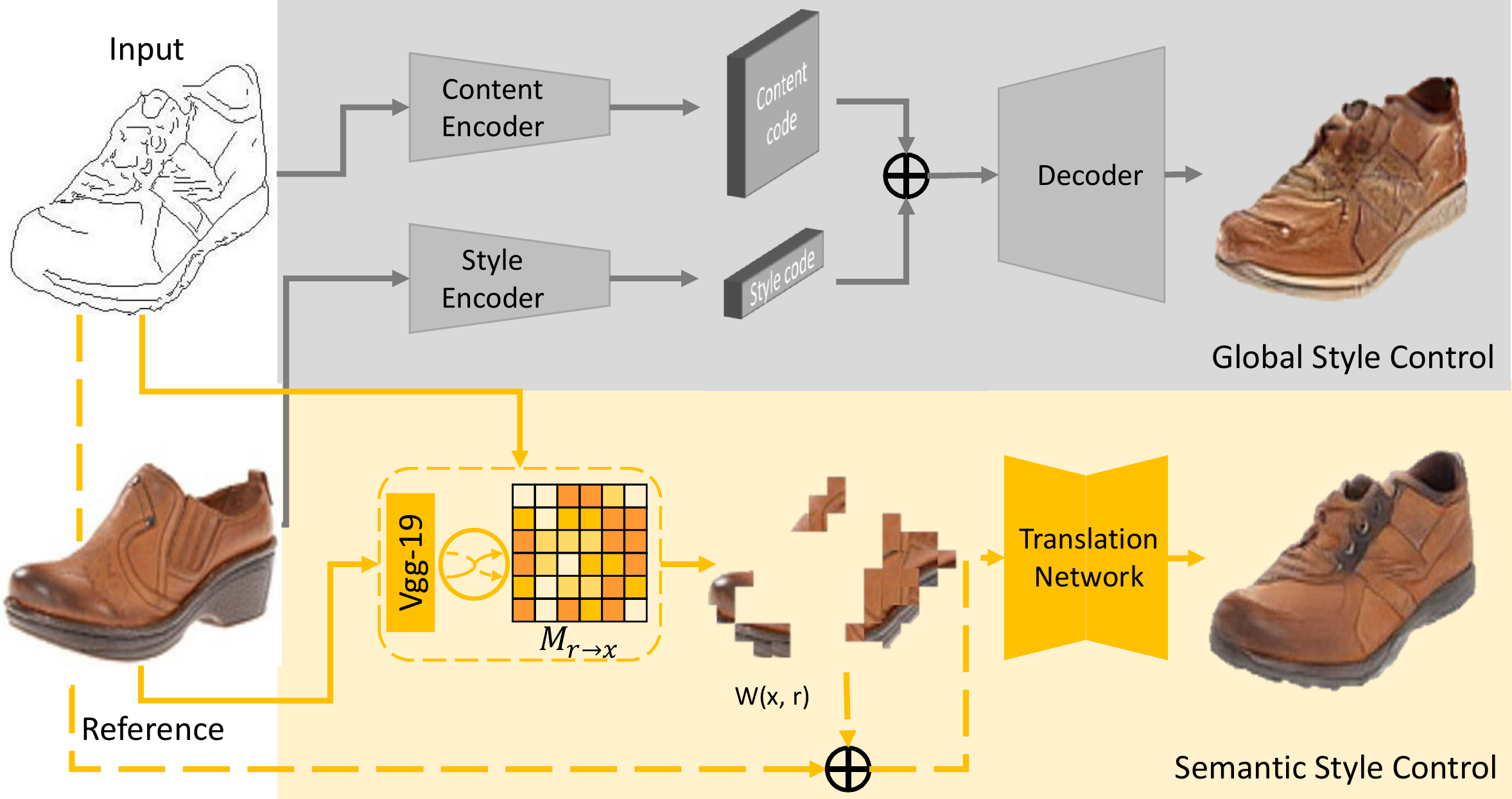}
    \caption{\small{The global style control can merely generate images with overall similarities to the reference, but our semantic style control preserves the characteristic of the reference with semantic matching. It can be observed that the color, the styles of the shoe sole and leather materials can be preserved in this example while the shoe generated by the conventional method can only inherit the general color of the reference. The correspondence map $M_{r\rightarrow x}$ and the auxiliary image $W(x, r)$ will be introduced with details in the section III.}}
    \label{fig:mechanism}
\end{figure}

The pioneering work Pix2pix\cite{pix2pix} designed a single-modal network as a general-purpose solution to the I2I problems, which is then followed by the CycleGAN\cite{cyclegan}. According to the CycleGAN, with cycle consistency, unpaired data can also be used as raw materials to form an outstanding I2I model. However, the I2I problem is in essence of high diversity that it is possible for one image in the source domain to have multiple counterparts in the target domain. Therefore researchers proposed multi-modal networks that can generate various images corresponding to one input. Recent progress shows that lots of the multi-modal networks are driven by noise vectors. Since noise vectors cannot provide specific guidance, a large number of experiments are then required to achieve desired results. Besides, modal collapse can frequently occur to this type of network due to that generator may ignore the additional noise in training\cite{noiseignore1}. Some multi-modal networks guided by images or attributes were then proposed \cite{diit,munit}. A common approach is to learn a low-dimensional style latent code of the reference image and then use it to reconstruct the output image; we name it as "Global Style Control" shown in Fig.\ref{fig:mechanism}. This method can generate results similar to the reference at an overall level, but fail to guarantee semantic similarity with details. To our knowledge, existing methods cannot perform semantically local control automatically in the I2I translation. \par
An ideal solution is to train a model that can not only learn mapping between two visual domains but also absorb semantic information from the reference images. This problem can be described in a more general way as a semantic exemplar-based multi-modal I2I problem. Given two sets of images in different domains $\mathcal{X}\subset\mathbb{R}^{H\times W\times C}$, $\mathcal{Y}\subset \mathbb{R}^{H\times W\times C}$ and a cluster of reference images $r\in \mathcal{Y}$, we train a model $G: (\mathcal{X}|r) \rightarrow \mathcal{Y}^r$. The output $\hat{y} = G(x|r)$ should be domain-wise indistinguishable from images $y\in \mathcal{Y}$ and contains the semantic affinity to the image $r\in \mathcal{Y}$. Here, $\mathcal{Y}^r \subset \mathcal{Y}$ refers to a sub-domain of $\mathcal{Y}$ that contains images of similar appearance style to $r$. This problem is, however, challenging for two main reasons: (1) lack of grouped data $(x, r, \hat{y})$. The output $\hat{y}$ should be in the target domain and contain visual attributes of the reference $r$; this kind of data is scarce and difficult to obtain. (2) Semantic incompatibility. The correspondence is first established between the input and the reference, since they may not have the identical semantic attributes, mismatches can happen during this process. Hence, the quality of the result rests heavily on the choice of the reference. \par

In this work, we present a self-supervised network that uses parts of reference images with the semantic affinity to the input image as guidance to ensure their semantic consistency. Compared with the aforementioned models with local control, our network only requires a reference image, which is of high flexibility and hence makes batch production possible in applications and manufacture.\par 

To achieve these goals, first, we design a pre-processing module that can build up semantic mapping between the input and the reference, generating a warped image as auxiliary data with visual attributes of the reference and preserving the structure of the input image. We explore the self-supervised method by constructing training data to conquer the difficulty of lack of grouped data. With paired data $\{x, y\}$ where $x\in \mathcal{X}$, $y\in \mathcal{Y}$, we warp $y$ to $x$ and obtain a warped image $y^{'}$. Then, we use $y^{'}$ as the reference for the input $x$ in the training stage. However, $y^{'}$ may have much more similarities with $x$ in comparison with the random reference used in real tests. Hence, we intentionally manipulate the warped images by shifting and adding noise, which aims to simulate mismatches and inconsistencies in the color, the shape, locations that $G$ will encounter in the test process. We then design multi-task architecture and non-local blocks to ameliorate performance of feature selection and guidance propagation.\par
Compared with conventional global style control mentioned in\cite{munit,spade}, the latent style code is extracted by an encoder in the form of a vector. Combined with the content structure, the final output can inherit overall style from the reference, shown in Fig.\ref{fig:mechanism} (the top part). However, this cannot achieve semantic control within a specific small region, for spatial information will be lost if the style code is stored in a vector. We propose the semantic-matching process to provide utmost semantic guidance for each sub-region. With the self-supervision, reliable guidance will be preserved since it contributes little to the reconstruction loss while the unreliable ones will be discarded or modified.\par
The results show that our model can be applied in various datasets and transfer reasonable results to the target domain with local semantic similarities to the reference images. We assessed the proposed method through extensive qualitative and quantitative evaluations, and our model presents competitive performance against the state-of-the-art methods. Additionally, a detailed ablation study in terms of both network architecture and loss functions demonstrates the effectiveness of our proposed model for multi-model I2I problems.  \par
Our major contributions are listed as follows:
\begin{itemize}
    \item We propose the first semantic example guided I2I translation solution, which can not only transfer images from one domain to another with style control based on the reference but also guarantee the output images to have semantic similarities with the reference. 
    \item To solve the lack of grouped data, we design a data configuration procedure, which then allows us to perform the translation in the manner of self-supervision. 
    \item We build up a new I2I model, including a semantic-matching block and a translation network. Non-local layers and multi-task architecture are adopted to achieve higher image quality
\end{itemize}

\section{Related Works}
\textbf{Style Transfer} Style transfer in a degree inspires the research work of I2I translation as it assumes images can be separated into content and style latent spaces. Gatys et al. \cite{gatys} first employed the CNN as a powerful tool for image content and style separation. Although it is very difficult to define the boundary for content and style, according to the experimental results of their work, the produced images look like the target style and meanwhile preserve their own dominant structures. Also, Elad et al. \cite{st_via_ts} proposed a style transfer algorithm to get stylized images closer in quality to the CNN ones as an extension of the traditional non-CNN methods. Later, Liu et al. proposed SuperBIG \cite{psbrush}, which emphasized styles related to a series of visual effects and created a photo stylistic brush based on Superpixel Bi Partite graph. Furthermore, novel methods like component analysis \cite{st_comp_analy} and structure preservation \cite{spst} are also used in style transfer to achieve visually appealing results.\par
\textbf{Single-modal I2I translation} Recent attempts for I2I translation are built on the Pix2pix \cite{pix2pix} model, which uses a conditional generative adversarial model to learn mapping between two visual domains. It is then followed by CycleGAN \cite{cyclegan}. Cycle consistency is proposed as the core component of their objectives, which provides adequate constraint to support training with unpaired data. The related work \cite{unit} proposed by Liu et al. assumes that images in different domains can be mapped to a shared latent space with the same latent code. BranchGAN \cite{branchgan} proposed a new structure with a shared encoder and dual decoders to capture the cross-domain distribution and generate images in both domains without any paired data. Moreover, Chen et al.\cite{qawI2I} explicitly employed a quality-aware loss at the domain level to improve the quality of the unsupervised I2I framework.\par
\textbf{Multi-modal I2I translation} However, those frameworks are single-modal networks that can merely output one image at a time, which is of low efficiency considering training time. To tackle this problem, in several research works \cite{discreet1,discreet2,pnn}, methods that can generate a discrete number of outputs with the same input are proposed. Later, Zhu et al. proposed BicycleGAN \cite{bicyclegan}, which can achieve many-to-many I2I translation with unlimited outputs. MUNIT \cite{munit} put forward by Huang et al. is also an outstanding multi-modal framework trained with unpaired data. Also, inspired by CycleGAN \cite{cyclegan}, Almahairi et al. proposed the Augmented CycleGAN \cite{augcycle} which can learn many-to-many mapping by cycling over the original domains augmented with auxiliary latent spaces.\par  
\textbf{I2I translation with global control} A significant limitation of most of these works is that generated outputs cannot be controlled explicitly. It was briefly mentioned in MUNIT\cite{munit} that the style code could be extracted from the reference image and be recombined with the content code of the input image to generate an image containing similar style as the reference. Instead of extracting the latent style code, Lee et al.\cite{diit} disentangled the latent spaces of x and y into a shared content space and an attribute space, respectively. Related works \cite{exemplar,distangle} also tried to manipulate latent space and encourage the model to generate multiple outputs based on provided references. However, separating and modifying the latent space can only ensure the similarity on the whole, but it cannot partially control the output image with such reference. In addition, \cite{asygan} is proposed to adapt the asymmetric domains by using an auxiliary variable to learn the additional data. However, it still focus on an overall style controlled by examples.\par
\textbf{I2I translation with local control} Few works focus on the local control in I2I translation; most of them require extract assistance from users. Zhang et al. \cite{localcontrolcolor} proposed a locally controlled coloration method requiring color indication from users at different locations. You et al.\cite{pirec} proposed a method to transfer the image from the edge domain with sparse color input. However, this method required carefully prepared guidance which provides detailed color distribution and position alignment with the input sketch.\par
\textbf{non-GAN based methods} In addition, there are also non-GAN based image synthesis works. Deep Image Analysis (DIA) proposed by Liao et al. \cite{dia} is a representative work of the iteration based traditional image synthesis method. The computation of a match map between features from an image pair can generate an output preserving the content structure of one image as well as style attributes from the other. Bansal et al. proposed PixelNN (PNN) \cite{pnn}, a simple nearest-neighbor (NN) approach that synthesizes high-frequency photo-realistic images from an "incomplete" signal, aiming to cope with the modal collapse issue introduced by GAN. PNN proposed another match method that uses the nearest-neighbor search method to learn the mapping between images and generates images preserving styles from different references. However, those methods may generate incorrect or invalid results once the corresponding map is inaccurate. Our proposed method shows robustness handling corresponding matching with noise; detailed comparison and discussion are shown in the following sections.

\section{the Semantic Example Guided I2I translation Network (SEGIN)}
\begin{figure*}[!ht]
    \centering
    \includegraphics[width=\textwidth]{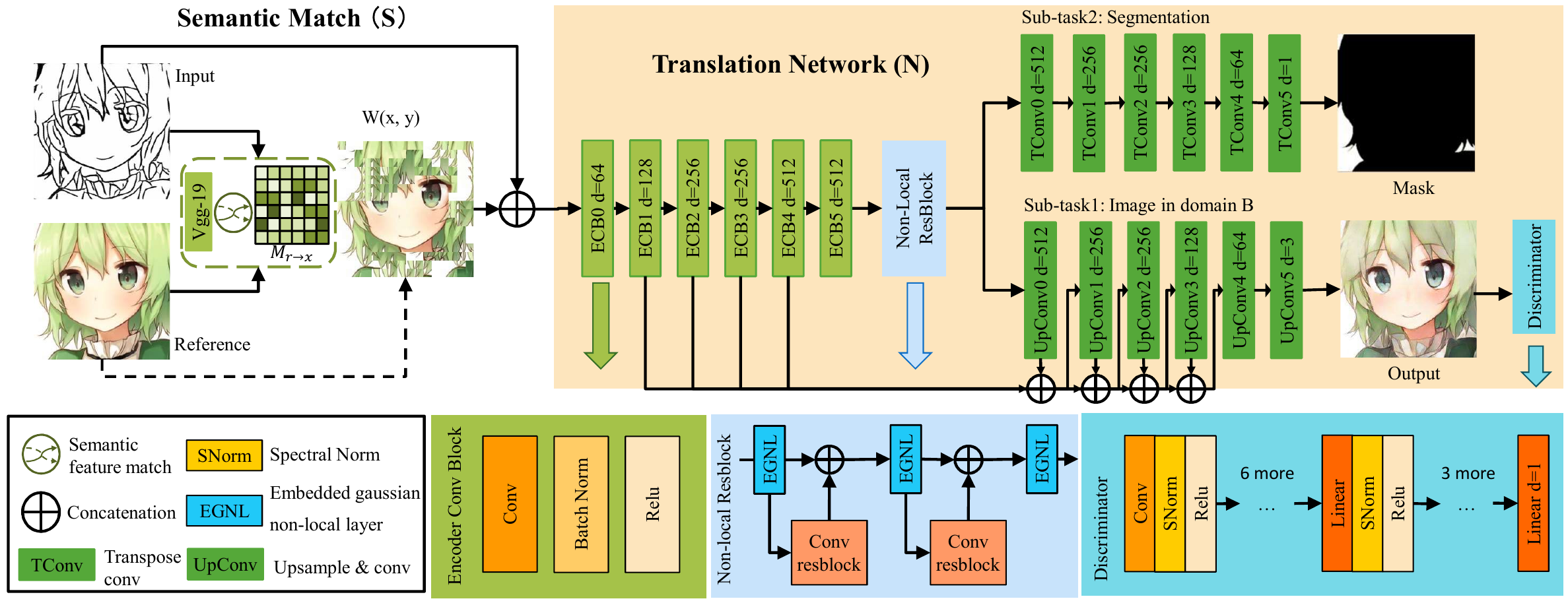}
    \caption{In the SEGIN model, shown at the left part is the semantic-match block, the mechanism of which is stated in later sub-sections. $\mathcal{W}(x, y)$ obtained by the semantic-match is then fed into the major network (shown on the right side) together with the input and the reference. There are two sub-tasks included in the translation network $N$, namely generating the images in target domains and the attention segmentation masks. Embedded Gaussian non-local layers are used in the middle Resblock while Spectral normalization layers \cite{sn} are adopted in the discriminator.}
    \label{fig:sys}
\end{figure*}

Our proposed I2I model $G$ consists of two parts, a preprocess module $S$ to match the $x$ and the $r$ semantically, and the translation network $N$ for I2I translation. Given an input image $x\in \mathcal{X}$ and a reference image $r\in \mathcal{Y}$, we first match image patches between $x$ and $r$ and generate an auxiliary image $\mathcal{W}(x, r)$, which contains superficial details of the reference and also maintains the basic structure of the input image. Both $x$ and $r$ are then fed into $N$ with auxiliary image $\mathcal{W}$, where the input image will be transferred to the target domain. Meanwhile, the semantic similarities to the reference image will be preserved. We provide an instantiation of the model $G: (\mathcal{X}|r) \rightarrow \mathcal{Y}^r$, in which the input image $x$, the final output $\hat{y}$ and the reference $r$ contain a set of sub-regions $\{x_1, x_2, ...,x_n\}$, $\{\hat{y}_1, \hat{y}_2, ...,\hat{y}_n\}$ and $\{r_1, r_2, ...,r_n\}$, respectively. Then, $\exists i, j\in\{1, 2, ..., n\}$ with $\hat{y}_i \sim x_i$ and $\hat{y}_i \sim r_j$, if the input image features around the sub-region $i$ are semantically related to those of the sub-region $j$ in the reference $r$, where "$A \sim B$" indicates that $A$ is similar to $B$ in pixel-wise. The architecture of our model the Semantic Example Guided I2I translation Network (SEGIN) is shown in Fig.\ref{fig:sys} and details will be presented and discussed as follows.\par
\subsection{Semantic Match}
\begin{figure}[!ht]
    \centering
    \includegraphics[width=\textwidth]{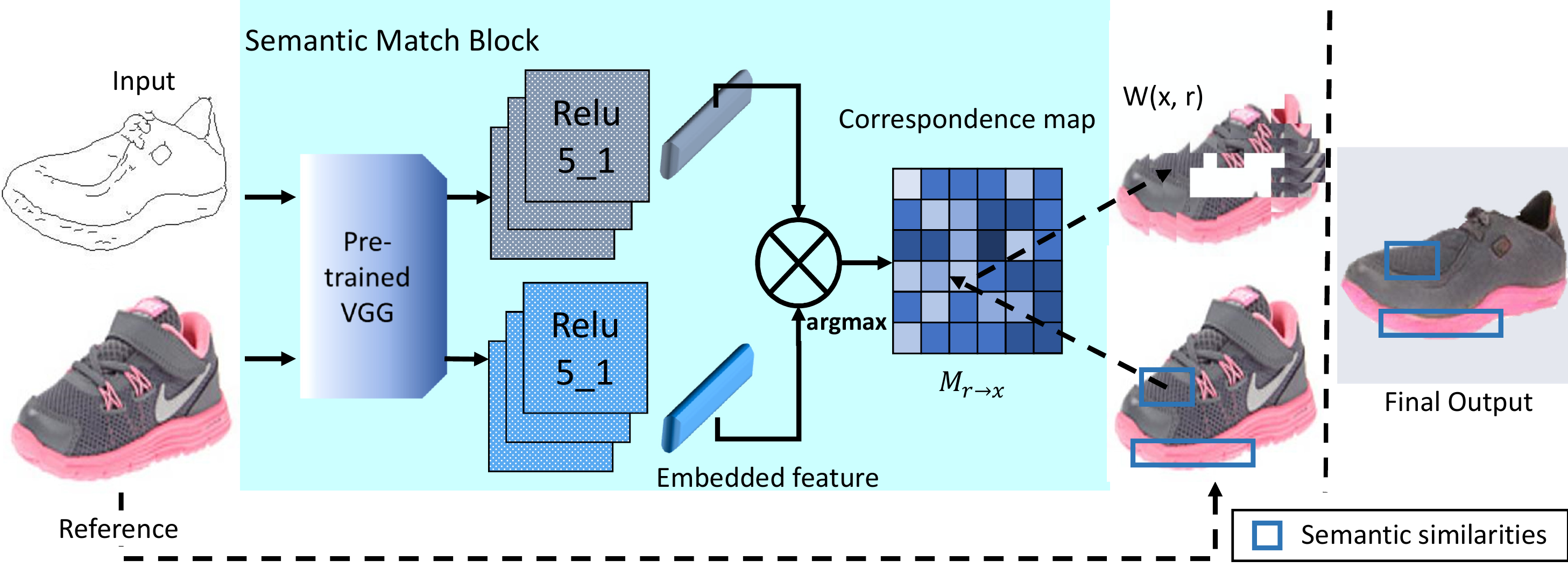}
    \caption{\small{Semantic match: Image features are first extracted from the input and the reference, respectively; cosine distance between the feature points is then calculated and the correspondence map $M_{r\rightarrow x}$ is generated. Reference image patches can be matched to the input based on the map, and those semantic and spatial information can finally be preserved in the output.}}
    \label{fig:sema}
\end{figure}

As shown in Fig.\ref{fig:sema}, image features of the input and the reference are first extracted by a pre-trained VGG-19 \cite{vgg}. Those features are then unfolded by the vectorization operation, followed by the generate corresponding map $M_{r\rightarrow x}$ between image features from different sources. We then construct the auxiliary image by adopting image patches from the reference in line with indication provided by the corresponding map. Spatially and semantically similar patches such as regions marked by the blue boxes on the reference shoes can still be kept, which will then be passed to the final outputs.\par
Define $\mathcal{\kappa}(\cdot)$ as a vectorization operation that transforms a matrix into a vector with row priority and let $\mathcal{\kappa}(\Phi(x))$ denotes a vector of all the image feature patches of $x$ with length of $n$, where the image features $\Phi(x)$ are extracted by VGG-19 \cite{vgg}. Our goal is to match the features between the reference and the input image. For each patch $\mathcal{\kappa}(\Phi(x))_i$ with size of $k\times k\times C$, where $C$ is the number of channels, we find its most similar feature patch $\mathcal{\kappa}(\Phi(r))_j$ in the reference by minimizing its cosine distance and building up a similarity map $\mathcal{\xi}$:
\begin{equation}
    \xi_{i} = \argmax_{j=1,2,...,n} \frac{\mathcal{\kappa}(\Phi(x))_i\cdot\mathcal{\kappa}(\Phi(r))_j}{|\mathcal{\kappa}(\Phi(x))_i|\cdot|\mathcal{\kappa}(\Phi(r))_j|}
\end{equation}

Image reconstruction methods mentioned in previous works \cite{diit,inpainting,munit} combine a compressed latent style code with the content, send them to the decoder and obtain the output. Semantic and spatial information may be washed away during this process \cite{spade}. We first construct the auxiliary image by adopting image patches in $r$ directly. Let $\mathcal{\kappa}(r)$ represent a list of image patches extracted from $r$ with the same length as $\mathcal{\kappa}(\Phi(r))$; the size of image patches $\mathcal{\kappa}(r)$ is $\delta k\times \delta k\times 3$ where $\delta$ is the scaling parameter and $\delta=size(r)/size(\Phi(r))$. Then, we define the reconstructed image as $\mathcal{W}(x, r)$ that
\begin{equation}
    \mathcal{\kappa}(\mathcal{W}(x,r))_i := \mathcal{\kappa}(r)_{\xi_i} 
\end{equation}
With a transform from the vector to the matrix $\mathcal{\kappa}^{-1}(\cdot)$, we can obtain the auxiliary image $W(x, r)$ that not only contains the superfacial details of the reference image with semantic similarities but also conserves the major structure of the input image. In this stage, we try to provide utmost semantically similar patches stored in $\mathcal{W}(x, r)$ as initial guidance for the local control.

\subsection{The Translation Network}
\subsubsection{\textbf{Data construction for the self-supervised model}}
According to the problem definition, the final output should belong to the target domain and preserve the semantically similar regions of $r$. This process requires conditioned mapping $G: (\mathcal{X}|r) \rightarrow \mathcal{Y}^r$, which differs from general supervised problems that always have ground truth for every output. \par
\begin{figure}[!ht]
    \centering
    \includegraphics[width=\textwidth]{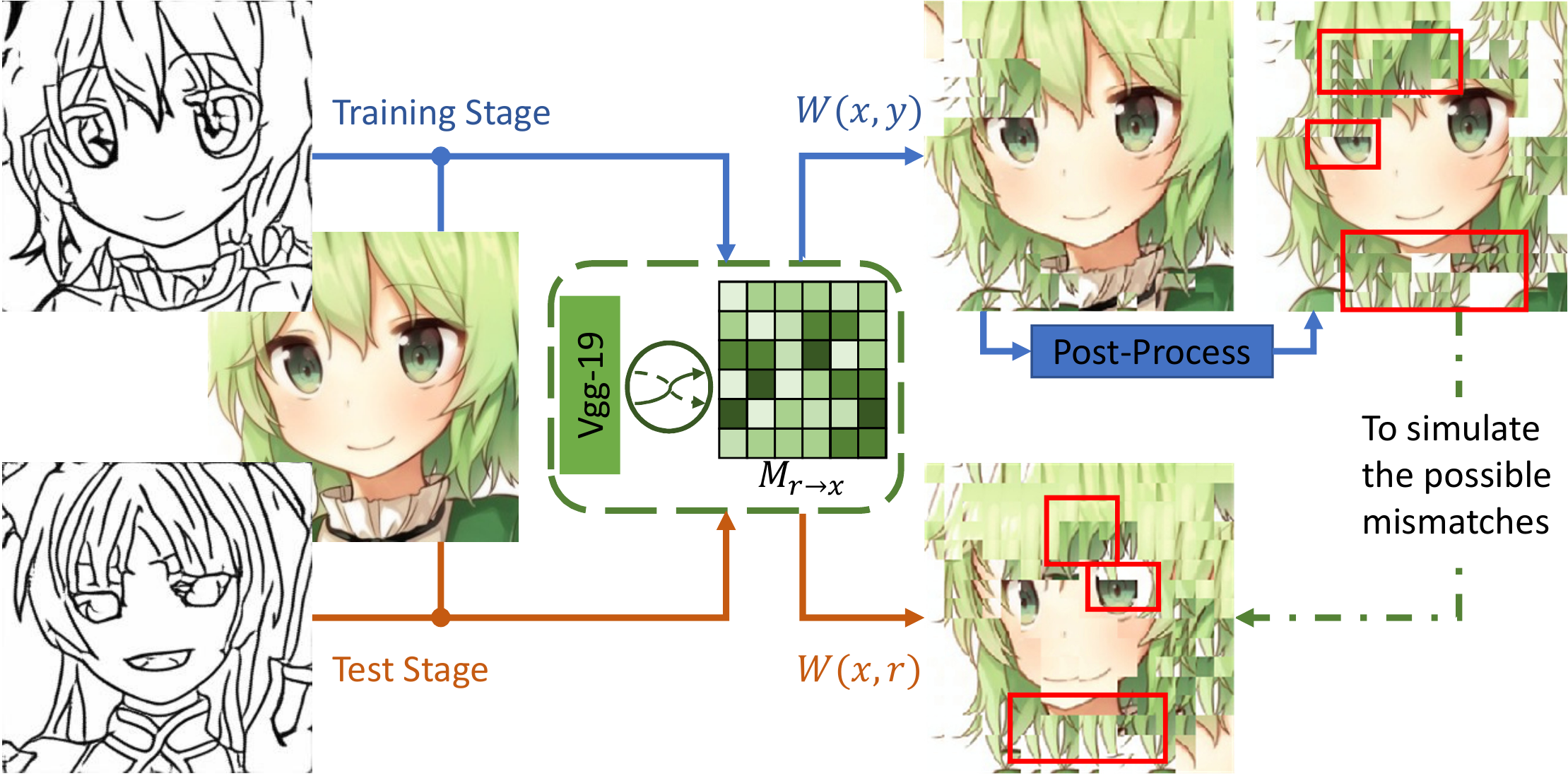}
    \caption{\small{Data construction and post-process}}
    \label{fig:postprocess}
\end{figure}

To settle this problem, we propose the self-supervised method. We use the paired data $y\in\{x, y\}$ as a fake reference of the input image $x$. Then, the required mapping can be rewritten as $G: (\mathcal{X}|\mathcal{Y})\rightarrow \mathcal{Y}^y$, and the desired output should be in keeping with the image $y$. However, this leads to another problem that $y$ has much more semantic similarities with $x$ than those with other random references, and in the test time, the auxiliary image $\mathcal{W}(x, r)$ would be much worse than the self-mapping result $\mathcal{W}(x, y)$ in terms of image quality as well as matching accuracy.  Inspired by \cite{semi}, we post-process the auxiliary image $\mathcal{W}(x, y)$, shown in Fig.\ref{fig:postprocess}. The post-processing includes patch shifting, repeating patterns, and random matches with a certain possibility, to simulate the mismatches and inaccuracy in $\mathcal{W}(x, r)$. Hence, the network can learn to fix the inconsistency between $\mathcal{W}(x, r)$ and the output in the aspects of locations, the color, and the shape.\par
This also explains the reasons why we separate the semantic match module from the training network. It would be too easy for the network to generate auxiliary images if it is an end-to-end fashion. With the self-supervised method, the input and the reference are paired data, and the network only needs to learn to copy the ground truth and make it as the auxiliary image $W(x, y)$. However, such method cannot generate valid auxiliary image $W(x, r)$ in the test stage when the reference has obvious structural difference from the input. With a separated process for generating the auxiliary images, followed by a post-processing, the semantic match can be functional in both training and test time.
\subsubsection{\textbf{Architecture}}
We use an auto-encoder structure for the generator $G$ of the translation network $\mathcal{N}$, which contains one encoder and two decoders for separated sub-tasks. The "U" net design \cite{unet} is adopted in our major decoder namely the decoder for generating images in the target domain. To enhance the propagation of local features, we apply a non-local block at the bottle neck layer of the generator. In our model, we use one discriminator for the major task, where spectral normalization \cite{sn} is applied.
In the initial experiment, we found that it was hard to output objects with smooth and complete surfaces, especially for the cases that foreground data distribution is vastly different from the background. Inspired by TextureGAN \cite{texturegan, style_controlled}, which used the segmentation directly as the input, we found that the segmentation can provide information similar to the attention mask mentioned in \cite{zebrapaper}. With that information, the network can focus on approximating the data distribution within the segmentation mask. However, it is not easy for users to obtain the segmentation masks in a real application if it is required as an input. Therefore, in our major network $N$, we utilize the multi-task structure with the major task (sub-task 1) generating images in the target domain, while the auxiliary task (sub-task 2) aims to generate segmentation distinguishing the foreground from the background of the ground truth images. It can therefore encourage the model to learn information about the attention area.\par
Moreover, there are cases that $\mathcal{W}(x, r)$ only comprises a few valid pixels in a small region. Owing to the limited receptive field used in the convolutional neural network, information cannot be propagated to a relatively large region. This issue is more serious when reliable guidance in $\mathcal{W}(x, r)$ is sparse, and the output image tends to be black or grayish since few data is generated based on the reference $r$. To deal with this problem, we adopt the non-local method \cite{nonlocal} by adding non-local layers into our translation network $N$. The general non-local operation is defined as:
\begin{equation}
    \mathbf{y}_i = \frac{1}{\mathcal{C}(x)}\sum_{\forall j}\mathcal{F}(\mathbf{x}_i, \mathbf{x}_j)g(\mathbf{x}_j)
    \label{nonlocal}
\end{equation}
where $i$ is the index of an output position whose response will be computed while $j$ is the index that enumerates all the possible positions. $\mathcal{F}$ is a pairwise function computing the affinity between $i$ and all $j$, and $g$ is a unary function that calculates a representation of the input signal at location $j$. In practice, the non-local blocks can be implemented in various formats by choosing disparate instantiations for $\mathcal{F}$. It is also mentioned in \cite{nonlocal} that the non-local models are not sensitive to the selection of specific functions, and therefore in our model we employ the embedded Gaussian function as $\mathcal{F}$ shown in Eq.\ref{embededgaussian}.
\begin{equation}
    \mathcal{F}(\mathbf{x}_i, \mathbf{x}_j)=e^{\theta(\mathbf{x}_i)^T\Phi(\mathbf{x}_j)}
    \label{embededgaussian}
\end{equation}
The non-local layer can also provide hints for the mismatched regions, as long as two regions $i, j$ possess similar content features. Style of the region with reliable guidance can also be propagated to the mismatched region, and related ablation studies are shown in Experiment.
\subsubsection{\textbf{Losses}}
The multi-task network is governed by the following objectives. 

\textbf{Self-reconstruction Loss} The target of this image generating task, as mentioned before, is a self-supervised process whose output is encouraged to be as similar as possible to the reference which is the ground truth in the training stage.
\begin{equation}
    \mathcal{L}_{recon}(N) = \mathbb{E}_{x, y, \mathcal{W}}[||y-N(x, \mathcal{W}(x,y))||_1]
    \label{eqn:l1}
\end{equation}

\textbf{Feature reconstruction Loss} Apart from the self-reconstruction loss, we also add the feature reconstruction loss. We extract image feature layers ($relu3\_2$) from VGG-19 \cite{vgg} to ensure the high-level similarities between the two images. It is shown in our experiments that the feature reconstruction loss can help to accelerate the image formation at the initial stage,
\begin{equation}
    \mathcal{L}_F(N)=\mathbb{E}_{x, y, \mathcal{W}}[||\Phi(N(x,\mathcal{W}(x, y)))-\Phi(y)||_1]
    \label{eqn:perceptual}
\end{equation}
where $\Phi(.)$ indicates the feature extraction.

\textbf{Adversarial Loss} We employ an adversarial loss to encourage the results of $N$ to be like the real samples from domain $\mathcal{Y}$. The loss is defined as:
\begin{equation}
    \begin{split}
    \mathcal{L}_{GAN}(N, D) &= \mathbb{E}_{x,y,W}[logD(x, y)]\\ 
    &+ \mathbb{E}_{x, \mathcal{W}}[log(1-D(N(x, \mathcal{W}(x, y))))] 
    \end{split}
   \label{eqn:gan}    
\end{equation}
where the discriminator $D$ endeavors to discriminate between the real samples from domain  $\mathcal{Y}$ and the generated samples $N(x, \mathcal{W}(x, y))$. Different from non-GAN based methods \cite{dia, pnn}. Our network can learn the distribution of a large dataset and thus is able to deal with improper matches in $W(x, r)$. It makes the proposed method robust to generate reasonable results even with auxiliary images containing unreliable guidance.

\textbf{Total Variation Loss} To make the output $\hat{y}$ smoother and more consistent, we also add a total variation loss as an objective.
\begin{equation}
   \mathcal{L}_{tv}(\hat{y}) = \sum_{i,j}\sqrt{|\hat{y}_{i+1, j} - \hat{y}_{i, j}|^2 + |\hat{y}_{i, j+1} - \hat{y}_{i, j}|^2}
\end{equation}

The second sub-task is to predict the segmentation map of the output image, serving as the attention mask. The attention loss consists of two parts. First, we qualify the segmentation by adding L1 norm regularization, shown in Eq.\ref{eqn:seg} 
\begin{equation}
    \mathcal{L}_{seg}(N_s) =\mathbb{E}_{x}[||N_s(x) - y_{seg}||_1]   
    \label{eqn:seg}
\end{equation}
where $y_{seg}$ denotes the segmentation ground truth.

We also want to add more shared information between the two tasks. Therefore, we propose the segmentation-based attention loss which is shown in Eq.\ref{eqn:acc_seg}.
\begin{equation}
    \begin{split}
    \mathcal{L}_{segAtt}(N, N_s) &=\mathbb{E}_{x, y,\mathcal{W}}[||N_s(x)\otimes N(x,\mathcal{W}(x,y))\\
    &- y_{seg}\otimes y||_1]   
    \end{split}
    \label{eqn:acc_seg}    
\end{equation}
where $\otimes$ represents the pixel-wise product. We use the output segmentation as an attention mask, and conduct element-wise product between the mask and the generated image, which is then qualified by an element-wise product of the real segmentation mask and the ground truth image.\par

The full objective is shown in Eq.\ref{eqn:full}
\begin{equation}
    \begin{split}
    \mathcal{L}_{total} &= w_1\mathcal{L}_{recon} + w_2\mathcal{L}_{F} + w_3\mathcal{L}_{GAN} \\
    &+ w_4\mathcal{L}_{seg} + w_5\mathcal{L}_{segAtt} + w_6\mathcal{L}_{tv}
    \end{split}
    \label{eqn:full}
\end{equation}
where $w_i$ is weight for each objective, obtained in accordance with the most optimal results across all the datasets tested in our work. Experiments on multiple datasets show that image qualities of the outputs of these two tasks can be enhanced mutually.

\section{Experiments}
\begin{figure*}[ht]
   \centering
\begin{subfigure}{1.8cm}
\includegraphics[width=1.8cm]{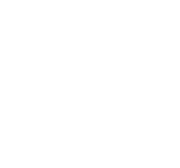}
\end{subfigure}
\begin{subfigure}{1.8cm}
\includegraphics[width=1.8cm]{fig/white.png}
\end{subfigure}
\begin{subfigure}{1.8cm}
\includegraphics[trim=0cm 0cm 0cm 1cm,clip=true,width=1.8cm]{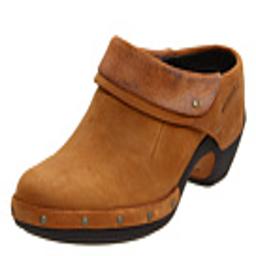}\vspace{-0.1cm}\caption*{\small{Ref 1}}
\end{subfigure}
\begin{subfigure}{1.8cm}
\includegraphics[trim=0cm 0cm 0cm 1cm,clip=true,width=1.8cm]{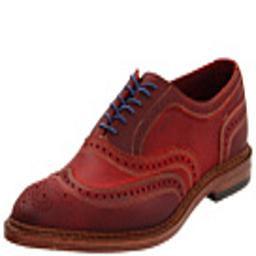}\vspace{-0.1cm}\caption*{\small{Ref 2}}
\end{subfigure}
\begin{subfigure}{1.8cm}
\includegraphics[trim=0cm 0cm 0cm 1cm,clip=true,width=1.8cm]{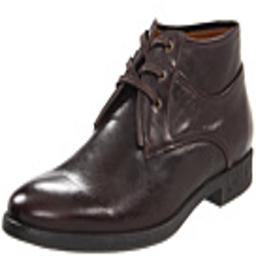}\vspace{-0.1cm}\caption*{\small{Ref 3}}
\end{subfigure}
\begin{subfigure}{1.8cm}
\includegraphics[trim=0cm 0cm 0cm 1cm,clip=true,width=1.8cm]{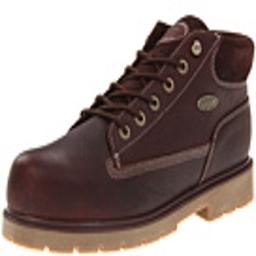}\vspace{-0.1cm}\caption*{\small{Ref 4}}
\end{subfigure}
\begin{subfigure}{1.8cm}
\includegraphics[trim=0cm 0cm 0cm 1cm,clip=true,width=1.8cm]{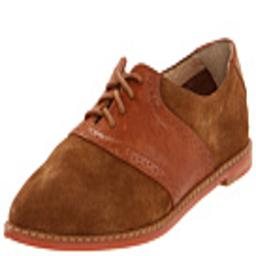}\vspace{-0.1cm}\caption*{\small{Ref 5}}
\end{subfigure}
\begin{subfigure}{1.8cm}
\includegraphics[trim=0cm 0cm 0cm 1cm,clip=true,width=1.8cm]{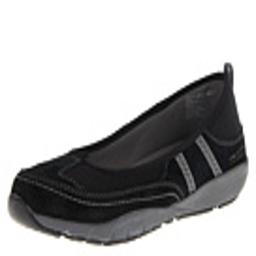}\vspace{-0.1cm}\caption*{\small{Ref 6}}
\end{subfigure}

\includegraphics[trim=0cm 0cm 0cm 1cm,clip=true,width=1.8cm]{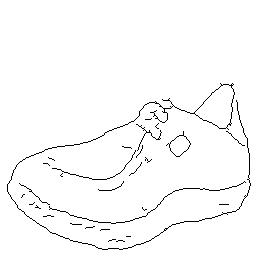}
\includegraphics[trim=0cm 0cm 0cm 1cm,clip=true,width=1.8cm]{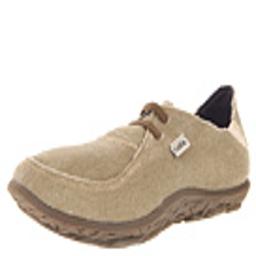}
\includegraphics[trim=0cm 0cm 0cm 1cm,clip=true,width=1.8cm]{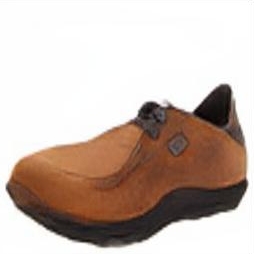}
\includegraphics[trim=0cm 0cm 0cm 1cm,clip=true,width=1.8cm]{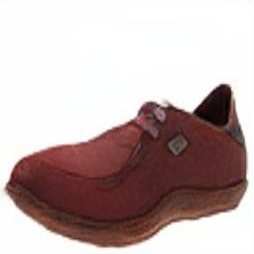}
\includegraphics[trim=0cm 0cm 0cm 1cm,clip=true,width=1.8cm]{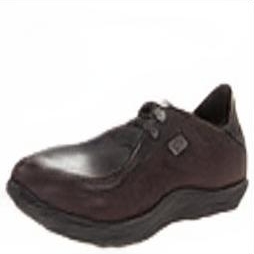}
\includegraphics[trim=0cm 0cm 0cm 1cm,clip=true,width=1.8cm]{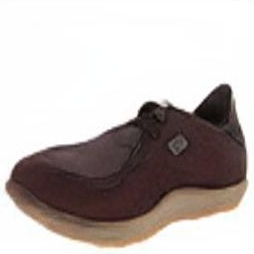}
\includegraphics[trim=0cm 0cm 0cm 1cm,clip=true,width=1.8cm]{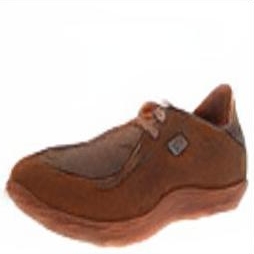}
\includegraphics[trim=0cm 0cm 0cm 1cm,clip=true,width=1.8cm]{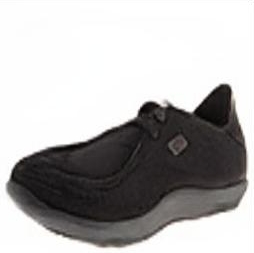}

\begin{subfigure}{1.8cm}
\includegraphics[trim=0cm 0cm 0cm 1cm,clip=true,width=1.8cm]{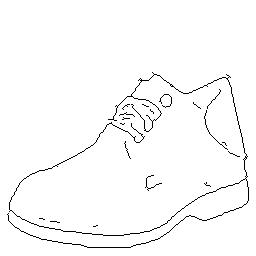}\vspace{-0.1cm}\caption*{\small{Input}}
\end{subfigure}
\begin{subfigure}{1.8cm}
\includegraphics[trim=0cm 0cm 0cm 1cm,clip=true,width=1.8cm]{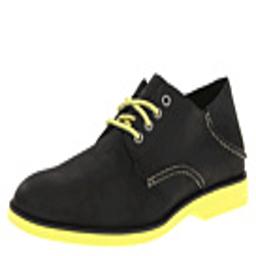}\vspace{-0.1cm}\caption*{\small{GT}}
\end{subfigure}
\begin{subfigure}{1.8cm}
\includegraphics[trim=0cm 0cm 0cm 1cm,clip=true,width=1.8cm]{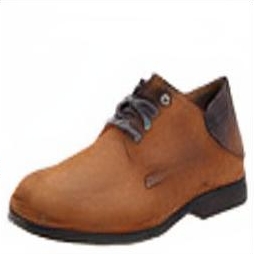}\vspace{-0.1cm}\caption*{\small{Result 1}}
\end{subfigure}
\begin{subfigure}{1.8cm}
\includegraphics[trim=0cm 0cm 0cm 1cm,clip=true,width=1.8cm]{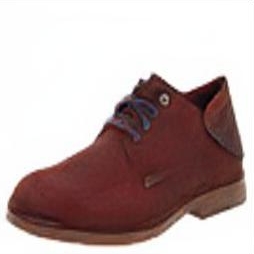}\vspace{-0.1cm}\caption*{\small{Result 2}}
\end{subfigure}
\begin{subfigure}{1.8cm}
\includegraphics[trim=0cm 0cm 0cm 1cm,clip=true,width=1.8cm]{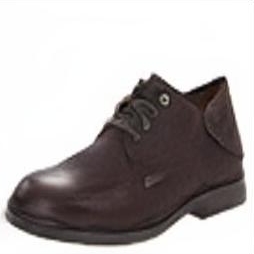}\vspace{-0.1cm}\caption*{\small{Result 3}}
\end{subfigure}
\begin{subfigure}{1.8cm}
\includegraphics[trim=0cm 0cm 0cm 1cm,clip=true,width=1.8cm]{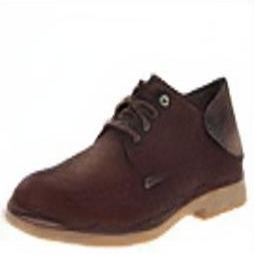}\vspace{-0.1cm}\caption*{\small{Result 4}}
\end{subfigure}
\begin{subfigure}{1.8cm}
\includegraphics[trim=0cm 0cm 0cm 1cm,clip=true,width=1.8cm]{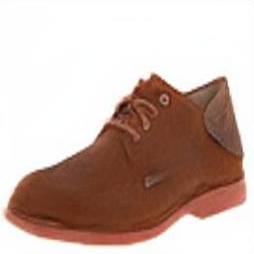}\vspace{-0.1cm}\caption*{\small{Result 5}}
\end{subfigure}
\begin{subfigure}{1.8cm}
\includegraphics[trim=0cm 0cm 0cm 1cm,clip=true,width=1.8cm]{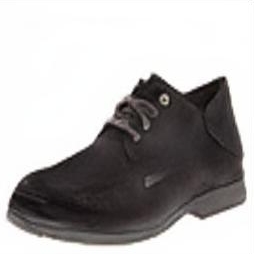}\vspace{-0.1cm}\caption*{\small{Result 6}}
\end{subfigure}

\caption{\small{Results generated by the SEGIN (our model) are various and natural. The results show semantic similarities to the given reference. Accordingly, Results 1-6 show that general styles of the reference can be inherited independently in specific locations; specifically, Result 3 (leather shoe) demonstrates that materials can also be transferred to the outputs.}}
\label{fig:shoes} 
\end{figure*}
\begin{figure*}[ht]
    \centering
    \captionsetup{size=small}
    \begin{subfigure}{1.6cm}
    \includegraphics[trim=0cm 0cm 0cm 0cm,clip=true,width=1.6cm]{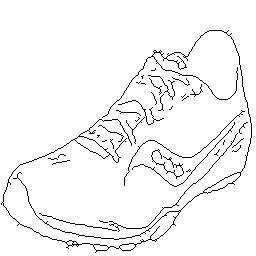}\vline
    \end{subfigure}
    \begin{subfigure}{1.6cm}
    \includegraphics[trim=0cm 0cm 0cm 0cm,clip=true,width=1.6cm]{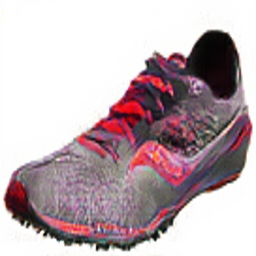}\vline
    \end{subfigure}
    \begin{subfigure}{1.6cm}
    \includegraphics[trim=0cm 0cm 0cm 0cm,clip=true,width=1.6cm]{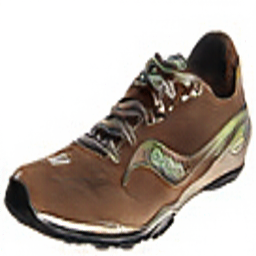}
    \end{subfigure}
    \begin{subfigure}{1.6cm}
    \includegraphics[trim=0cm 0cm 0cm 0cm,clip=true,width=1.6cm]{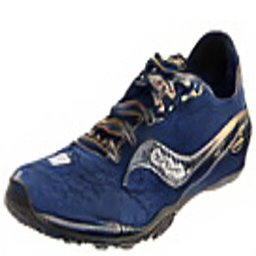}\vline
    \end{subfigure}
    \begin{subfigure}{1.6cm}
    \includegraphics[trim=0cm 0cm 0cm 0cm,clip=true,width=1.6cm]{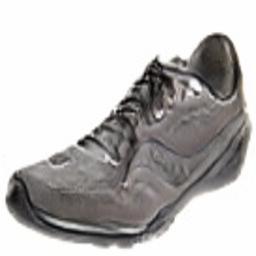}
    \end{subfigure}
    \begin{subfigure}{1.6cm}
    \frame{\includegraphics[trim=0cm 0cm 0cm 0cm,clip=true,width=1.6cm]{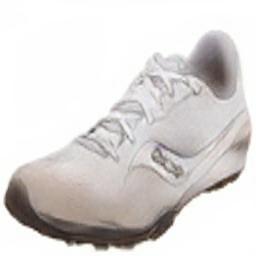}}
    \end{subfigure}
    \begin{subfigure}{1.6cm}
    \includegraphics[trim=0cm 0cm 0cm 0cm,clip=true,width=1.6cm]{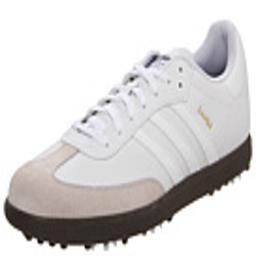}\vline
    \end{subfigure}
    \begin{subfigure}{1.6cm}
    \includegraphics[trim=0cm 0cm 0cm 0cm,clip=true,width=1.6cm]{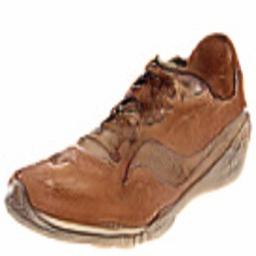}
    \end{subfigure}
    \begin{subfigure}{1.6cm}
    \frame{\includegraphics[trim=0cm 0cm 0cm 0cm,clip=true,width=1.6cm]{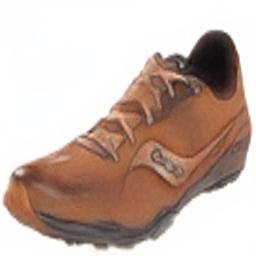}}
    \end{subfigure}
    \begin{subfigure}{1.6cm}
    \includegraphics[trim=0cm 0cm 0cm 0cm,clip=true,width=1.6cm]{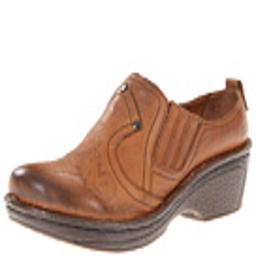}
    \end{subfigure}

    \begin{subfigure}{1.6cm}
    \includegraphics[trim=0cm 0cm 0cm 0cm,clip=true,width=1.6cm]{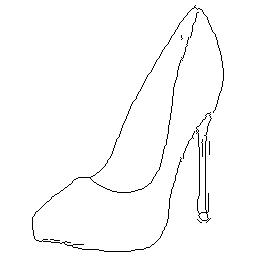}\vline
    \end{subfigure}
    \begin{subfigure}{1.6cm}
    \includegraphics[trim=0cm 0cm 0cm 0cm,clip=true,width=1.6cm]{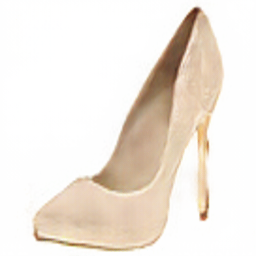}\vline
    \end{subfigure}
    \begin{subfigure}{1.6cm}
    \includegraphics[trim=0cm 0cm 0cm 0cm,clip=true,width=1.6cm]{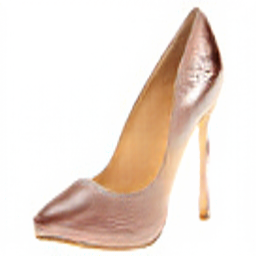}
    \end{subfigure}
    \begin{subfigure}{1.6cm}
    \includegraphics[trim=0cm 0cm 0cm 0cm,clip=true,width=1.6cm]{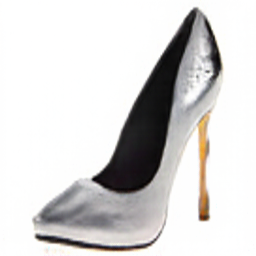}\vline
    \end{subfigure}
    \begin{subfigure}{1.6cm}
    \includegraphics[trim=0cm 0cm 0cm 0cm,clip=true,width=1.6cm]{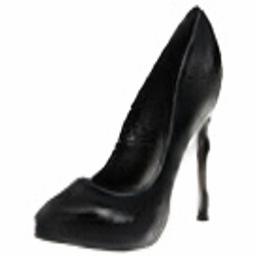}
    \end{subfigure}
    \begin{subfigure}{1.6cm}
    \frame{\includegraphics[trim=0cm 0cm 0cm 0cm,clip=true,width=1.6cm]{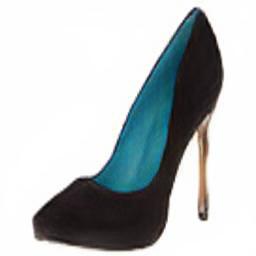}}
    \end{subfigure}
    \begin{subfigure}{1.6cm}
    \includegraphics[trim=0cm 0cm 0cm 0cm,clip=true,width=1.6cm]{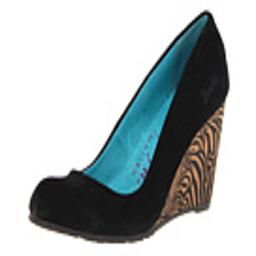}\vline
    \end{subfigure}
    \begin{subfigure}{1.6cm}
    \includegraphics[trim=0cm 0cm 0cm 0cm,clip=true,width=1.6cm]{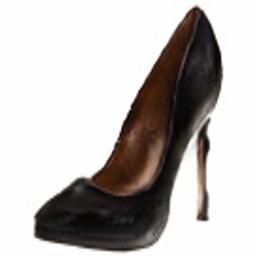}
    \end{subfigure}
    \begin{subfigure}{1.6cm}
    \frame{\includegraphics[trim=0cm 0cm 0cm 0cm,clip=true,width=1.6cm]{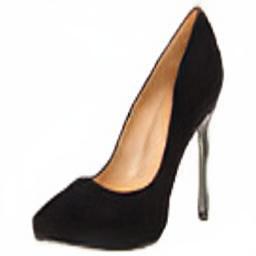}} 
    \end{subfigure}
    \begin{subfigure}{1.6cm}
    \includegraphics[trim=0cm 0cm 0cm 0cm,clip=true,width=1.6cm]{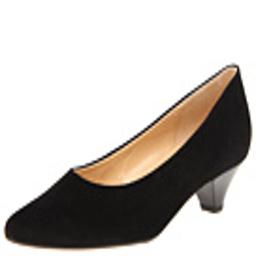}
    \end{subfigure}
    
    \begin{subfigure}{1.6cm}
    \includegraphics[trim=0cm 0cm 0cm 0cm,clip=true,width=1.6cm]{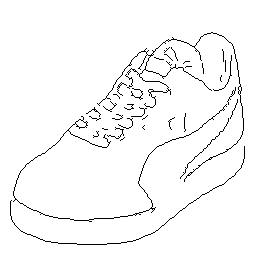}\vline
    \end{subfigure}
    \begin{subfigure}{1.6cm}
    \includegraphics[trim=0cm 0cm 0cm 0cm,clip=true,width=1.6cm]{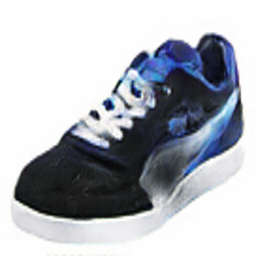}\vline
    \end{subfigure}
    \begin{subfigure}{1.6cm}
    \includegraphics[trim=0cm 0cm 0cm 0cm,clip=true,width=1.6cm]{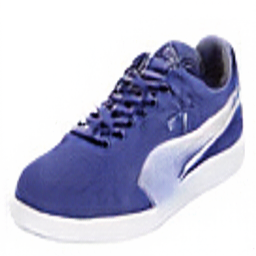}
    \end{subfigure}
    \begin{subfigure}{1.6cm}
    \includegraphics[trim=0cm 0cm 0cm 0cm,clip=true,width=1.6cm]{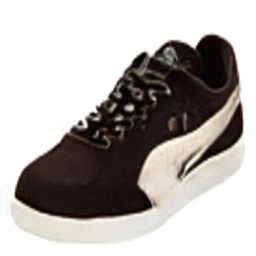}\vline
    \end{subfigure}
    \begin{subfigure}{1.6cm}
    \includegraphics[trim=0cm 0cm 0cm 0cm,clip=true,width=1.6cm]{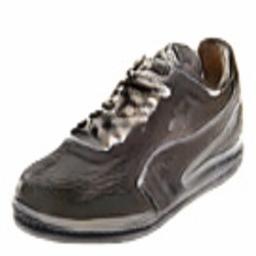}
    \end{subfigure}
    \begin{subfigure}{1.6cm}
    \frame{\includegraphics[trim=0cm 0cm 0cm 0cm,clip=true,width=1.6cm]{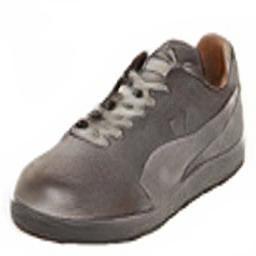}}
    \end{subfigure}
    \begin{subfigure}{1.6cm}
    \includegraphics[trim=0cm 0cm 0cm 0cm,clip=true,width=1.6cm]{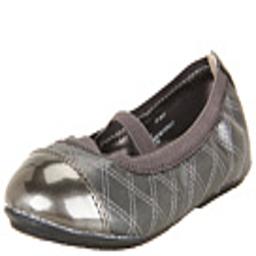}\vline
    \end{subfigure}
    \begin{subfigure}{1.6cm}
    \includegraphics[trim=0cm 0cm 0cm 0cm,clip=true,width=1.6cm]{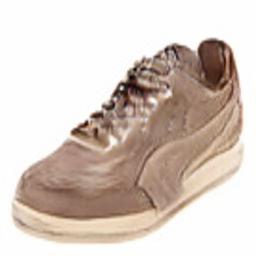}
    \end{subfigure}
    \begin{subfigure}{1.6cm}
    \frame{\includegraphics[trim=0cm 0cm 0cm 0cm,clip=true,width=1.6cm]{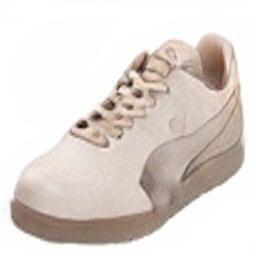}}
    \end{subfigure}
    \begin{subfigure}{1.6cm}
    \includegraphics[trim=0cm 0cm 0cm 0cm,clip=true,width=1.6cm]{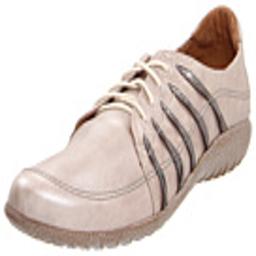}
    \end{subfigure}

     \begin{subfigure}{1.6cm}
    \includegraphics[trim=0cm 0cm 0cm 0cm,clip=true,width=1.6cm]{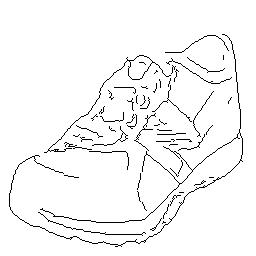}\vline\caption*{\small{Input}}
    \end{subfigure}
    \begin{subfigure}{1.6cm}
    \includegraphics[trim=0cm 0cm 0cm 0cm,clip=true,width=1.6cm]{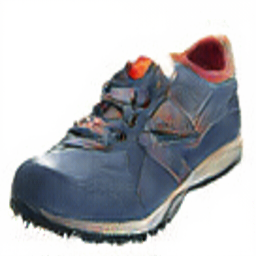}\vline\caption*{\small{Pix2pix}}
    \end{subfigure}
    \begin{subfigure}{1.6cm}
    \includegraphics[trim=0cm 0cm 0cm 0cm,clip=true,width=1.6cm]{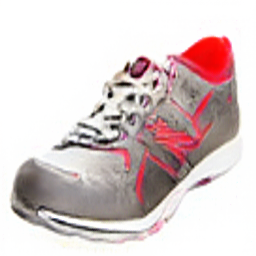}\caption*{\small{BiCycle 1}}
    \end{subfigure}
    \begin{subfigure}{1.6cm}
    \includegraphics[trim=0cm 0cm 0cm 0cm,clip=true,width=1.6cm]{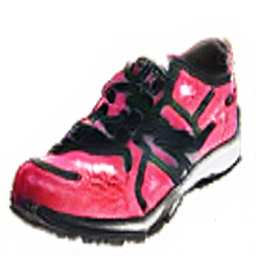}\vline\caption*{\small{BiCycle 2}}
    \end{subfigure}
    \begin{subfigure}{1.6cm}
    \includegraphics[trim=0cm 0cm 0cm 0cm,clip=true,width=1.6cm]{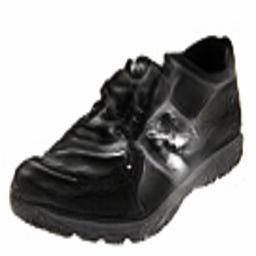}\caption*{\small{MUNIT 1}}
    \end{subfigure}
    \begin{subfigure}{1.6cm}
    \frame{\includegraphics[trim=0cm 0cm 0cm 0cm,clip=true,width=1.6cm]{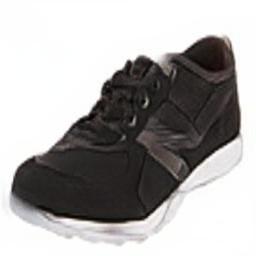}}\caption*{\small{Ours 1}}
    \end{subfigure}
    \begin{subfigure}{1.6cm}
    \includegraphics[trim=0cm 0cm 0cm 0cm,clip=true,width=1.6cm]{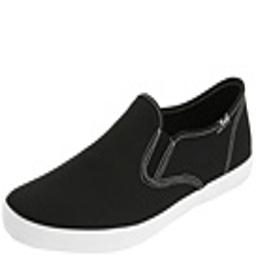}\vline\caption*{\small{Ref 1}}
    \end{subfigure}
    \begin{subfigure}{1.6cm}
    \includegraphics[trim=0cm 0cm 0cm 0cm,clip=true,width=1.6cm]{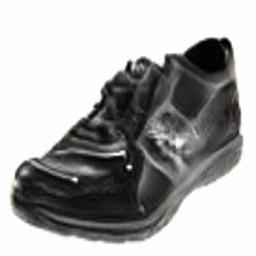}\caption*{\small{MUNIT 2}}
    \end{subfigure}
    \begin{subfigure}{1.6cm}
    \frame{\includegraphics[trim=0cm 0cm 0cm 0cm,clip=true,width=1.6cm]{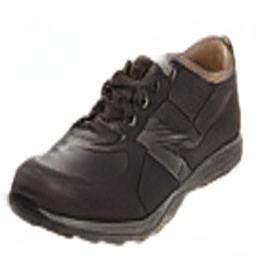}}\caption*{\small{Ours 2}}
    \end{subfigure}
    \begin{subfigure}{1.6cm}
    \includegraphics[trim=0cm 0cm 0cm 0cm,clip=true,width=1.6cm]{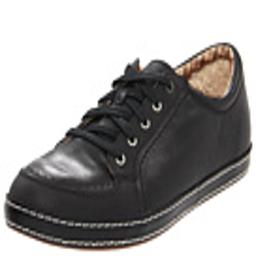}\caption*{\small{Ref 2}}
    \end{subfigure}
    \caption{\small{Compared to Pix2pix and the BiCycleGAN, our model can generate images that are not only with varieties but also controlled by the reference; we also compare our model with MUNIT in terms of similarity to the reference. We use the same references $Ref_i$ for both MUNIT and our models, and we observe that: (1) our results are of higher semantically similarity to the reference (e.g. in the $2^{nd}$ row, MUNIT failed to generate high heel shoes with blue or brown lining); (2) with the similar reference, modal collapse happens in the MUNIT model.(MUNIT 1 and MUNIT 2 in the last row)}}
    \label{fig:comp}
\end{figure*}
\textbf{Baseline} We evaluate several prior I2I models, including Pix2pix \cite{pix2pix}, BiCycleGAN \cite{bicyclegan} and MUNIT \cite{munit}, and set them as the baseline. Specifically, it is stated in MUNIT \cite{munit} that a single image can be used as a style reference in image generation. We also set it as the baseline of the image-guided I2I model and conduct related experiments for comparison. Besides, we also compared our results with those non-GAN based works, and found that DIA \cite{dia}, PNN \cite{pnn} are typical baselines standing for traditional iteration based image synthesis and parametric image synthesis without GAN. We first compare our results directly with results of those methods. Since those methods have similarities to the semantic match part designed in our framework, we also conduct the ablation study that replaces our semantic match parts with DIA \cite{dia} and PNN \cite{pnn} to show it is also possible to use different match methods introducing the semantic guidance, and our architecture is robust and of high flexibility with different auxiliary images.\par
\textbf{Dataset} We test our method on several typical I2I translation datasets, covering edges2photots \cite{edge1, edge2}, Danbooru2018 \cite{danbooru2018} and face colorization on CelebA \cite{celeba}. For Danbooru \cite{danbooru2018}, we combine the Hed \cite{hed} with sketch simplification \cite{sketchimprove} method to extract the sketch of images; those sketches are then used as training and test data in domain A. This sketch generation method can provide the better sketch compared with the original sketches obtained by HED\cite{hed} used in edges2photos dataset, and some results based on such improved sketches are shown in the Appendix. As for the segmentation used in the training stages, we simply pick colored regions as foreground and the remaining parts as the background for Danbooru2018 \cite{danbooru2018} and CelebA \cite{celeba}. The segmentation of edges2photos can be found in the work TextureGAN \cite{texturegan}. All the models are trained with 256x256 images. \par 
\textbf{General settings} Our major network contains an encoder for the reference and the input image, two decoders for the target image and the segmentation. We use 6 combined convolution blocks with the ReLu activation layer and batch normalization layers for the encoder and add 3 non-local layers in total at the bottleneck. Residual blocks with two 3x3 convolutional blocks are used in between non-local layers. To avoid checkerboard effects mentioned in the previous work \cite{checkerboard}, we use a combination of the up-sample and the convolution to replace deconvolution layers in the decoders of sub-task 1. Instead of using batch normalization, we adopt spectral normalization \cite{sn} as training stabilization techniques in our discriminator to achieve better image quality. We applied different learning rate for the generator ($2e^{-4}$) and the discriminator ($1.3e^{-5}$), and weights of different objectives are: $w_1=w_4=100.0$, $w_2=5e^{-6}$, $w_4=1.0$, $w_5=w_6=10.0$. Parameters are kept the same for all the experiments in this work unless specific modification is mentioned later. For datasets containing more than 50K images (sketch2shoes \cite{edge1}, sketch2bags \cite{edge2}, and celebA \cite{celeba}), we use a batch size of 5 and around 20 epochs to achieve satisfactory and stable results. With two NVIDIA 2080Ti GPUs, it takes about 20 hours to finish the training. For smaller datasets, 12-epoch with training time around 10 hours should be enough. For all the datasets, the test time required by the translation network is around 0.03s while the semantic match costs about 1s for a single image.\par

\begin{figure*}[!ht]
    \centering
    \includegraphics[width=1.8cm]{fig/white.png}
    \includegraphics[width=1.8cm]{fig/white.png}
    \begin{subfigure}{1.8cm}
    \includegraphics[trim=0cm 0cm 0cm 0cm,clip=true,width=1.8cm]{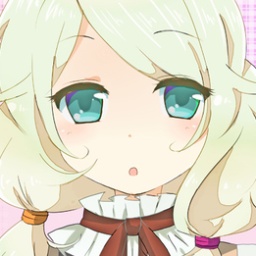}\vspace{-0.1cm}\caption*{\small{Ref 1}}
    \end{subfigure}
    \begin{subfigure}{1.8cm}
    \includegraphics[trim=0cm 0cm 0cm 0cm,clip=true,width=1.8cm]{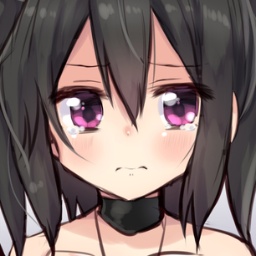}\vspace{-0.1cm}\caption*{\small{Ref 2}}
    \end{subfigure}
    \begin{subfigure}{1.8cm}
    \includegraphics[trim=0cm 0cm 0cm 0cm,clip=true,width=1.8cm]{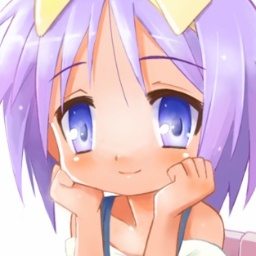}\vspace{-0.1cm}\caption*{\small{Ref 3}}
    \end{subfigure}
    \begin{subfigure}{1.8cm}
    \includegraphics[trim=0cm 0cm 0cm 0cm,clip=true,width=1.8cm]{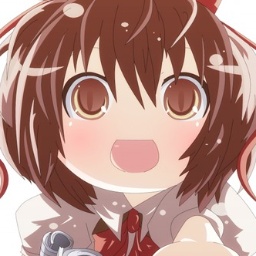}\vspace{-0.1cm}\caption*{\small{Ref 4}}
    \end{subfigure}
    \begin{subfigure}{1.8cm}
    \includegraphics[trim=0cm 0cm 0cm 0cm,clip=true,width=1.8cm]{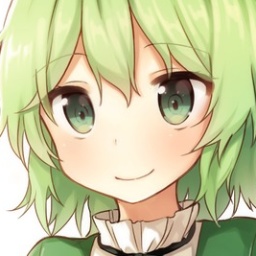}\vspace{-0.1cm}\caption*{\small{Ref 5}}
    \end{subfigure}
    \begin{subfigure}{1.8cm}
    \includegraphics[trim=0cm 0cm 0cm 0cm,clip=true,width=1.8cm]{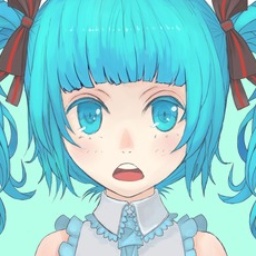}\vspace{-0.1cm}\caption*{\small{Ref 6}}
    \end{subfigure}
    
    \includegraphics[trim=0cm 0cm 0cm 0cm,clip=true,width=1.8cm]{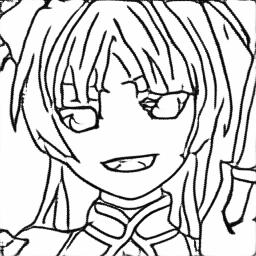}
    \includegraphics[trim=0cm 0cm 0cm 0cm,clip=true,width=1.8cm]{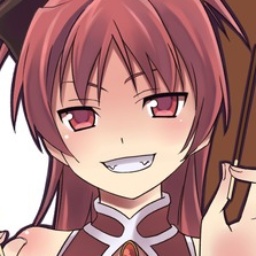}
    \includegraphics[trim=0cm 0cm 0cm 0cm,clip=true,width=1.8cm]{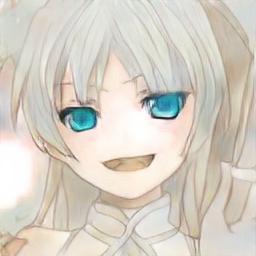}
    \includegraphics[trim=0cm 0cm 0cm 0cm,clip=true,width=1.8cm]{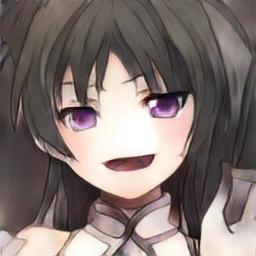}
    \includegraphics[trim=0cm 0cm 0cm 0cm,clip=true,width=1.8cm]{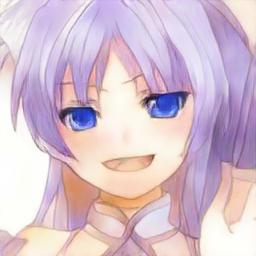}
    \includegraphics[trim=0cm 0cm 0cm 0cm,clip=true,width=1.8cm]{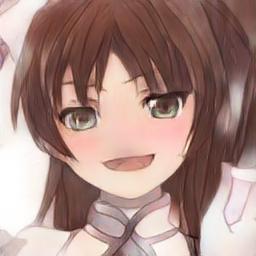}
    \includegraphics[trim=0cm 0cm 0cm 0cm,clip=true,width=1.8cm]{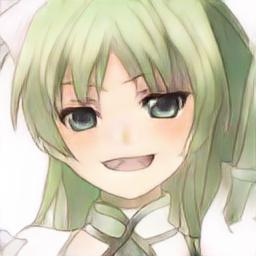}
    \includegraphics[trim=0cm 0cm 0cm 0cm,clip=true,width=1.8cm]{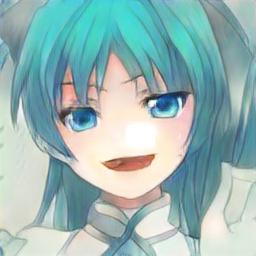}
    
    \begin{subfigure}{1.8cm}
    \includegraphics[trim=0cm 0cm 0cm 0cm,clip=true,width=1.8cm]{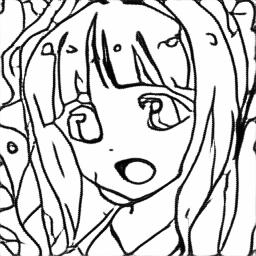}\vspace{-0.1cm}\caption*{\small{Input}}
    \end{subfigure}
    \begin{subfigure}{1.8cm}
    \includegraphics[trim=0cm 0cm 0cm 0cm,clip=true,width=1.8cm]{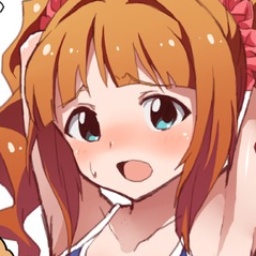}\vspace{-0.1cm}\caption*{\small{GT}}
    \end{subfigure}
    \begin{subfigure}{1.8cm}
    \includegraphics[trim=0cm 0cm 0cm 0cm,clip=true,width=1.8cm]{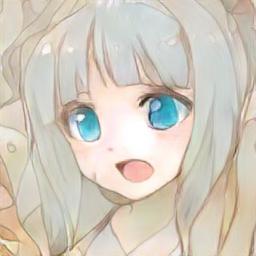}\vspace{-0.1cm}\caption*{\small{Result 1}}
    \end{subfigure}
    \begin{subfigure}{1.8cm}
    \includegraphics[trim=0cm 0cm 0cm 0cm,clip=true,width=1.8cm]{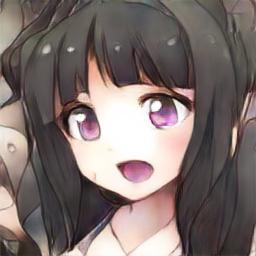}\vspace{-0.1cm}\caption*{\small{Result 2}}
    \end{subfigure}
    \begin{subfigure}{1.8cm}
    \includegraphics[trim=0cm 0cm 0cm 0cm,clip=true,width=1.8cm]{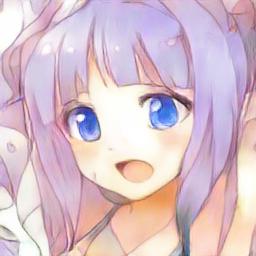}\vspace{-0.1cm}\caption*{\small{Result 3}} 
    \end{subfigure}
    \begin{subfigure}{1.8cm}
    \includegraphics[trim=0cm 0cm 0cm 0cm,clip=true,width=1.8cm]{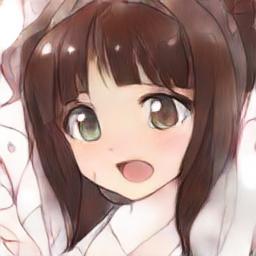}\vspace{-0.1cm}\caption*{\small{Result 4}} 
    \end{subfigure}
    \begin{subfigure}{1.8cm}
    \includegraphics[trim=0cm 0cm 0cm 0cm,clip=true,width=1.8cm]{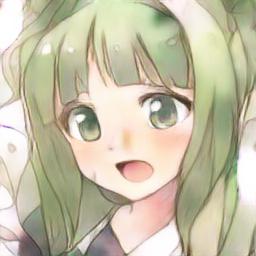}\vspace{-0.1cm}\caption*{\small{Result 5}}
    \end{subfigure}
    \begin{subfigure}{1.8cm}
    \includegraphics[trim=0cm 0cm 0cm 0cm,clip=true,width=1.8cm]{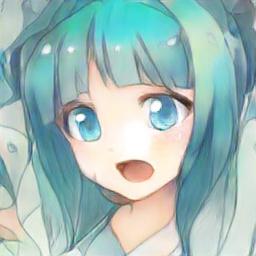}\vspace{-0.1cm}\caption*{\small{Result 6}} 
    \end{subfigure}
    \caption{\small{We test our model on Danbooru\cite{danbooru2018} anime dataset for the sketch $\rightarrow$ the anime task. It can be observed that many semantic characteristics of the reference, including eyes, hairs, the blush, etc. are preserved in the generated results.}}
    \label{fig:ani}
   \end{figure*} 

\subsection{Qualitative Evaluation}
\begin{figure*}[!ht]
    \centering
    \includegraphics[width=1.8cm]{fig/white.png}
    \includegraphics[width=1.8cm]{fig/white.png}
    \begin{subfigure}{1.8cm}
    \includegraphics[trim=0cm 0cm 0cm 0cm,clip=true,width=1.8cm]{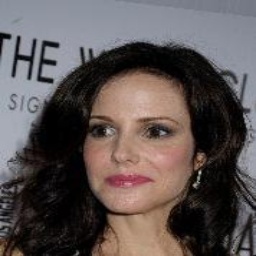}\vspace{-0.1cm}\caption*{\small{Ref 1}} 
    \end{subfigure}
    \begin{subfigure}{1.8cm}
    \includegraphics[trim=0cm 0cm 0cm 0cm,clip=true,width=1.8cm]{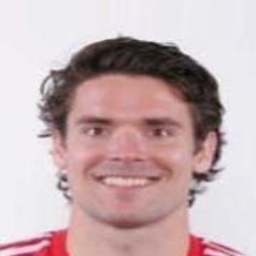}\vspace{-0.1cm}\caption*{\small{Ref 2}} 
    \end{subfigure}
    \begin{subfigure}{1.8cm}
    \includegraphics[trim=0cm 0cm 0cm 0cm,clip=true,width=1.8cm]{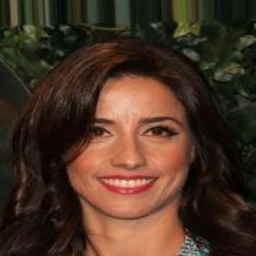}\vspace{-0.1cm}\caption*{\small{Ref 3}} 
    \end{subfigure}
    \begin{subfigure}{1.8cm}
    \includegraphics[trim=0cm 0cm 0cm 0cm,clip=true,width=1.8cm]{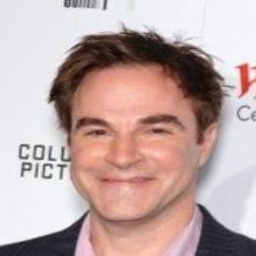}\vspace{-0.1cm}\caption*{\small{Ref 4}} 
    \end{subfigure}
    \begin{subfigure}{1.8cm}
    \includegraphics[trim=0cm 0cm 0cm 0cm,clip=true,width=1.8cm]{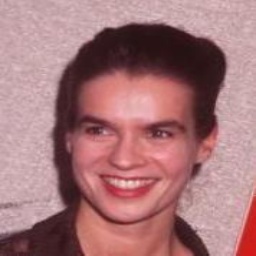}\vspace{-0.1cm}\caption*{\small{Ref 5}} 
    \end{subfigure}
    \begin{subfigure}{1.8cm}
    \includegraphics[trim=0cm 0cm 0cm 0cm,clip=true,width=1.8cm]{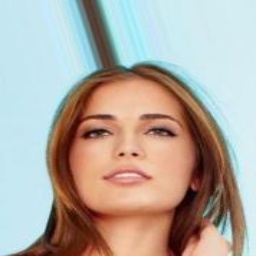}\vspace{-0.1cm}\caption*{\small{Ref 6}} 
    \end{subfigure}
    
    \includegraphics[trim=0cm 0cm 0cm 0cm,clip=true,width=1.8cm]{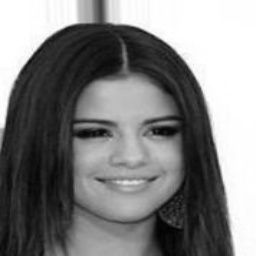}
    \includegraphics[trim=0cm 0cm 0cm 0cm,clip=true,width=1.8cm]{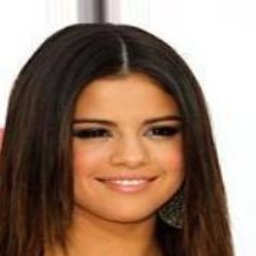}
    \includegraphics[trim=0cm 0cm 0cm 0cm,clip=true,width=1.8cm]{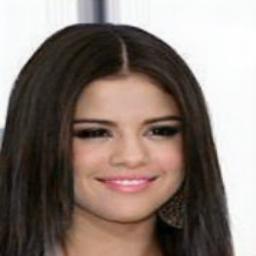}
    \includegraphics[trim=0cm 0cm 0cm 0cm,clip=true,width=1.8cm]{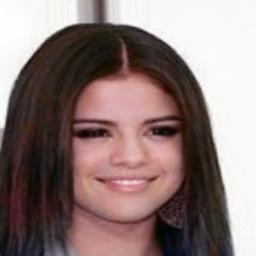}
    \includegraphics[trim=0cm 0cm 0cm 0cm,clip=true,width=1.8cm]{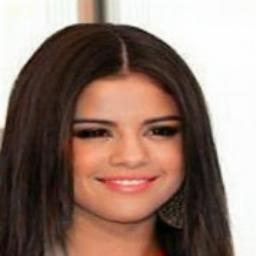}
    \includegraphics[trim=0cm 0cm 0cm 0cm,clip=true,width=1.8cm]{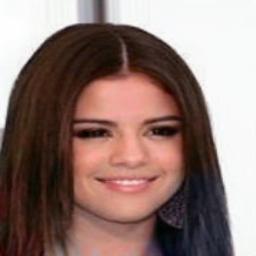}
    \includegraphics[trim=0cm 0cm 0cm 0cm,clip=true,width=1.8cm]{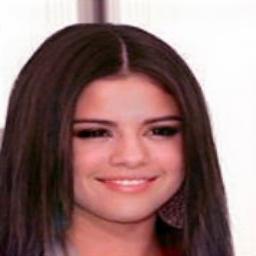}
    \includegraphics[trim=0cm 0cm 0cm 0cm,clip=true,width=1.8cm]{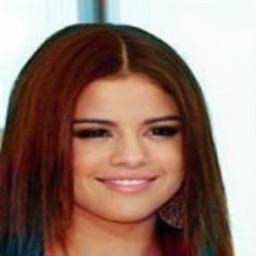}
    
    \begin{subfigure}{1.8cm}
    \includegraphics[trim=0cm 0cm 0cm 0cm,clip=true,width=1.8cm]{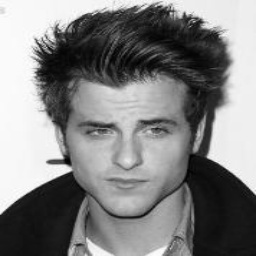}\vspace{-0.1cm}\caption*{\small{Input}}
    \end{subfigure}
    \begin{subfigure}{1.8cm}
    \includegraphics[trim=0cm 0cm 0cm 0cm,clip=true,width=1.8cm]{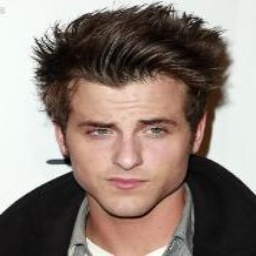}\vspace{-0.1cm}\caption*{\small{GT}}
    \end{subfigure}
    \begin{subfigure}{1.8cm}
    \includegraphics[trim=0cm 0cm 0cm 0cm,clip=true,width=1.8cm]{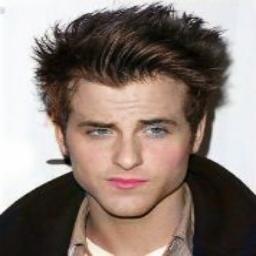}\vspace{-0.1cm}\caption*{\small{Result 1}}
    \end{subfigure}
    \begin{subfigure}{1.8cm}
    \includegraphics[trim=0cm 0cm 0cm 0cm,clip=true,width=1.8cm]{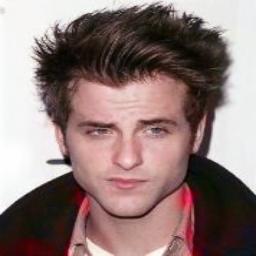}\vspace{-0.1cm}\caption*{\small{Result 2}}
    \end{subfigure}
    \begin{subfigure}{1.8cm}
    \includegraphics[trim=0cm 0cm 0cm 0cm,clip=true,width=1.8cm]{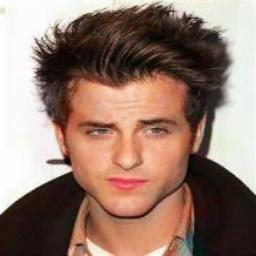}\vspace{-0.1cm}\caption*{\small{Result 3}}
    \end{subfigure}
    \begin{subfigure}{1.8cm}
    \includegraphics[trim=0cm 0cm 0cm 0cm,clip=true,width=1.8cm]{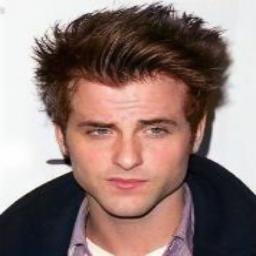}\vspace{-0.1cm}\caption*{\small{Result 4}}
    \end{subfigure}
    \begin{subfigure}{1.8cm}
    \includegraphics[trim=0cm 0cm 0cm 0cm,clip=true,width=1.8cm]{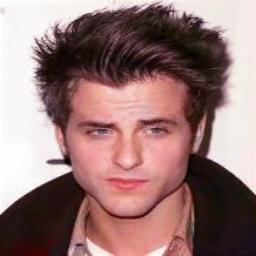}\vspace{-0.1cm}\caption*{\small{Result 5}}
    \end{subfigure}
    \begin{subfigure}{1.8cm}
    \includegraphics[trim=0cm 0cm 0cm 0cm,clip=true,width=1.8cm]{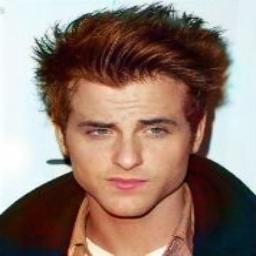}\vspace{-0.1cm}\caption*{\small{Result 6}} 
    \end{subfigure}
    \caption{We also test our model for face colorization on the CelebA\cite{celeba}. Generated images inherit the style from the reference semantically, only the color of facial elements like skip, lips, etc. are changed accordingly.}
    \label{fig:face}
\end{figure*}
We qualitatively compare our model SEGIN with the previously mentioned models, and the comparison results on the edge2shoes dataset are shown in Fig.\ref{fig:comp}. It can be observed that Pix2pix can only output a deterministic result for an input, while BiCycleGAN can output results with diversity, but the styles of the output cannot be controlled specifically. MUNIT adds image references to guide the output but obtains relatively blurry results with artifacts. Moreover, outputs of the MUNIT can simply preserve the style of the reference in general. We notice that the SEGIN presents similarities to the reference semantically; it can also maintain semantic details of the reference, including different color distribution at specific locations and object materials (i.e., leather, canvas, the suede surface, etc.). More results on the aforementioned dataset are shown in the Appendix.\par
Examples in different datasets are shown in Fig.\ref{fig:ani}-\ref{fig:face}, and we observe that the SEGIN is of diversity, reality  as well as semantic similarities to the corresponding reference. In the anime dataset, the SEGIN generates images with similar eyes, hairs as well as blushers to that of the reference. Furthermore, in the CelebA dataset, we observe that not only the skin color but also the makeup on the reference can be transferred to the outputs.
\begin{figure}[ht]
     \centering
     \begin{subfigure}{1.6cm}
    \includegraphics[trim=0cm 0cm 0cm 0cm,clip=true,width=1.6cm]{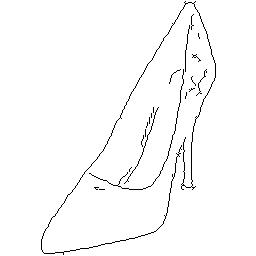}
    \end{subfigure}
     \begin{subfigure}{1.6cm}
    \includegraphics[trim=0cm 0cm 0cm 0cm,clip=true,width=1.6cm]{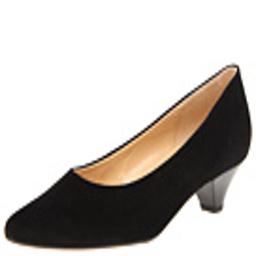}\vline
    \end{subfigure}
    \begin{subfigure}{1.6cm}
    \includegraphics[trim=0cm 0cm 0cm 0cm,clip=true,width=1.6cm]{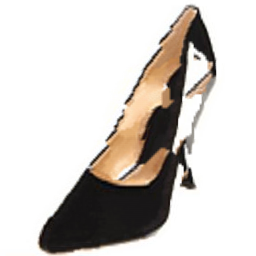}
    \end{subfigure}
    \begin{subfigure}{1.6cm}
    \includegraphics[trim=0cm 0cm 0cm 0cm,clip=true,width=1.6cm]{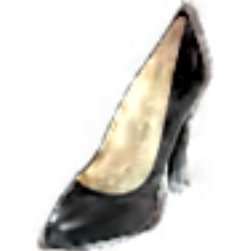}
    \end{subfigure}
    \begin{subfigure}{1.6cm}
    \frame{\includegraphics[trim=0cm 0cm 0cm 0cm,clip=true,width=1.6cm]{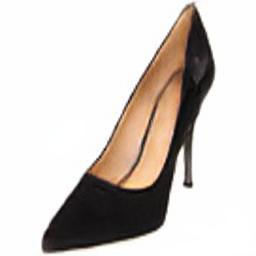}}
    \end{subfigure}
    
     \begin{subfigure}{1.6cm}
    \includegraphics[trim=0cm 0cm 0cm 0cm,clip=true,width=1.6cm]{fig/nonGAN/134_skg.jpg}
    \end{subfigure}
    \begin{subfigure}{1.6cm}
    \includegraphics[trim=0cm 0cm 0cm 0cm,clip=true,width=1.6cm]{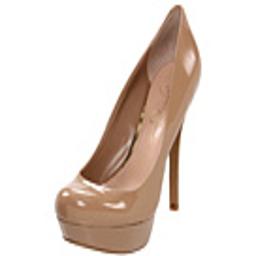}\vline
    \end{subfigure}
    \begin{subfigure}{1.6cm}
    \includegraphics[trim=0cm 0cm 0cm 0cm,clip=true,width=1.6cm]{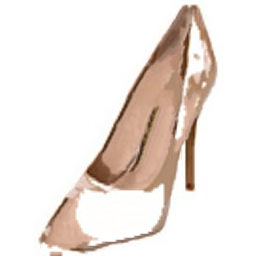}
    \end{subfigure}
    \begin{subfigure}{1.6cm}
    \includegraphics[trim=0cm 0cm 0cm 0cm,clip=true,width=1.6cm]{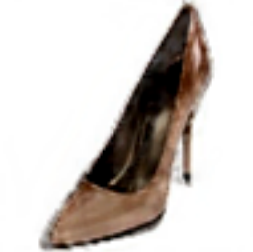}
    \end{subfigure}
    \begin{subfigure}{1.6cm}
    \frame{\includegraphics[trim=0cm 0cm 0cm 0cm,clip=true,width=1.6cm]{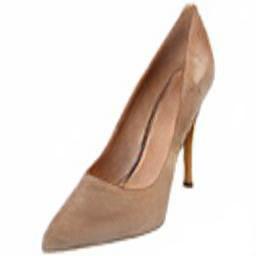}}
    \end{subfigure}
    
     \begin{subfigure}{1.6cm}
    \includegraphics[trim=0cm 0cm 0cm 0cm,clip=true,width=1.6cm]{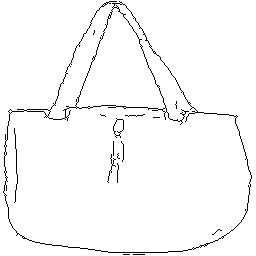}
    \end{subfigure}
    \begin{subfigure}{1.6cm}
    \includegraphics[trim=0cm 0cm 0cm 0cm,clip=true,width=1.6cm]{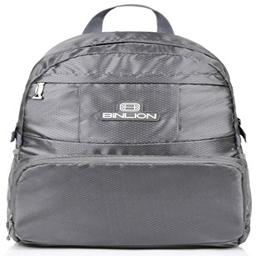}\vline
    \end{subfigure}
    \begin{subfigure}{1.6cm}
    \includegraphics[trim=0cm 0cm 0cm 0cm,clip=true,width=1.6cm]{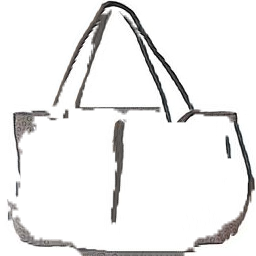}
    \end{subfigure}
    \begin{subfigure}{1.6cm}
    \includegraphics[trim=0cm 0cm 0cm 0cm,clip=true,width=1.6cm]{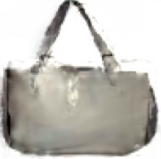}
    \end{subfigure}
    \begin{subfigure}{1.6cm}
    \frame{\includegraphics[trim=0cm 0cm 0cm 0cm,clip=true,width=1.6cm]{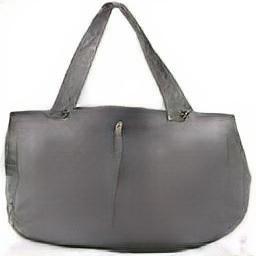}}
    \end{subfigure}

     \begin{subfigure}{1.6cm}
    \includegraphics[trim=0cm 0cm 0cm 0cm,clip=true,width=1.6cm]{fig/nonGAN/107_skg.jpg}\caption*{\small{Input}}
    \end{subfigure}   
    \begin{subfigure}{1.6cm}
    \includegraphics[trim=0cm 0cm 0cm 0cm,clip=true,width=1.6cm]{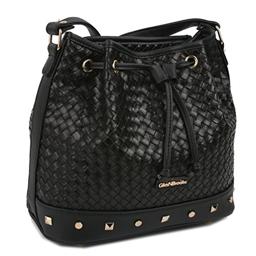}\vline\caption*{\small{Ref}}
    \end{subfigure}
    \begin{subfigure}{1.6cm}
    \includegraphics[trim=0cm 0cm 0cm 0cm,clip=true,width=1.6cm]{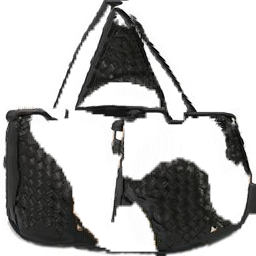}\caption*{\small{DIA}}
    \end{subfigure}
    \begin{subfigure}{1.6cm}
    \includegraphics[trim=0cm 0cm 0cm 0cm,clip=true,width=1.6cm]{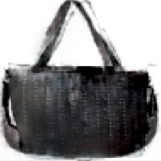}\caption*{\small{PNN}}
    \end{subfigure}
    \begin{subfigure}{1.6cm}
    \frame{\includegraphics[trim=0cm 0cm 0cm 0cm,clip=true,width=1.6cm]{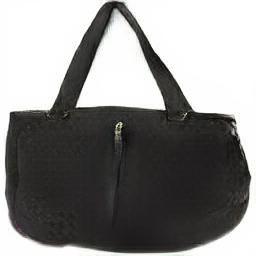}}\caption*{\small{Ours}}
    \end{subfigure}
     \caption{\small{Comparison with the non-GAN based image synthesis methods. It can be observed that results of DIA\cite{dia} can be invalid if there is no proper corresponding patches between the input and the reference while the results of PNN\cite{pnn} are quite blurry. GAN can learn to generate images in the target domain even with little hints and corresponding information.}}
     \label{fig:patchAblation}
 \end{figure}
We add the comparison between our work and the non-GAN based methods (i.e., DIA\cite{dia}, PNN\cite{pnn}), and results are shown in Fig.\ref{fig:patchAblation}. It is prominent that the results of PNN\cite{pnn} are quite blurry, especially at the image boundaries. While the output of DIA\cite{dia} show that results may be improper when it cannot find the correct corresponding patches in the reference, that is, those methods purely relying on matching output invalid (DIA \cite{dia}) or blurry (unnatural) results (PNN \cite{pnn}) when the correspondences are unreliable. In contrast, our GAN-based framework shows higher robustness, since it can learn to repair mismatched regions from a large dataset within a specific distribution. \par

\begin{figure*}[!ht]
     \centering
    \begin{subfigure}{1.6cm}
    \includegraphics[trim=0cm 0cm 0cm 0cm,clip=true,width=1.6cm]{fig/nonGAN/134_skg.jpg}
    \end{subfigure}
    \begin{subfigure}{1.6cm}
    \includegraphics[trim=0cm 0cm 0cm 0cm,clip=true,width=1.6cm]{fig/nonGAN/12_AB.jpg}\vline
    \end{subfigure}
    \begin{subfigure}{1.6cm}
    \includegraphics[trim=0cm 0cm 0cm 0cm,clip=true,width=1.6cm]{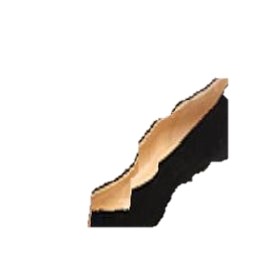}
    \end{subfigure}
    \begin{subfigure}{1.6cm}
    \includegraphics[trim=0cm 0cm 0cm 0cm,clip=true,width=1.6cm]{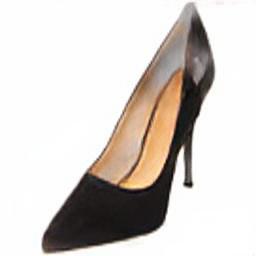}\vline
    \end{subfigure}
    \begin{subfigure}{1.6cm}
    \includegraphics[trim=0cm 0cm 0cm 0cm,clip=true,width=1.6cm]{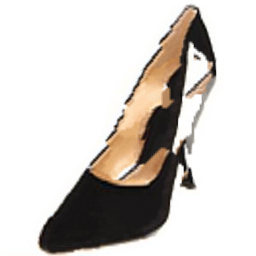}
    \end{subfigure}
    \begin{subfigure}{1.6cm}
    \includegraphics[trim=0cm 0cm 0cm 0cm,clip=true,width=1.6cm]{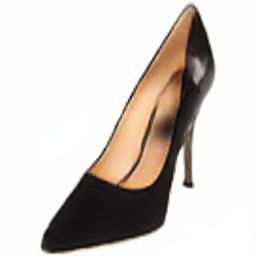}\vline
    \end{subfigure}
    \begin{subfigure}{1.6cm}
    \includegraphics[trim=0cm 0cm 0cm 0cm,clip=true,width=1.6cm]{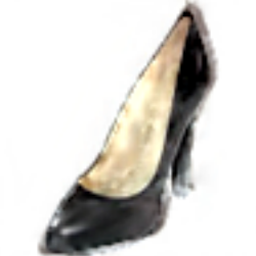}
    \end{subfigure}
    \begin{subfigure}{1.6cm}
    \includegraphics[trim=0cm 0cm 0cm 0cm,clip=true,width=1.6cm]{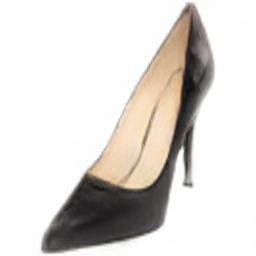}\vline
    \end{subfigure}
    \begin{subfigure}{1.6cm}
    \includegraphics[trim=0cm 0cm 0cm 0cm,clip=true,width=1.6cm]{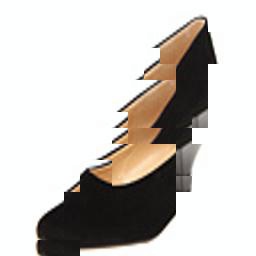}
    \end{subfigure}
    \begin{subfigure}{1.6cm}
    \frame{\includegraphics[trim=0cm 0cm 0cm 0cm,clip=true,width=1.6cm]{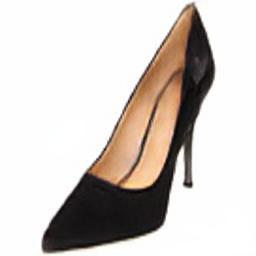}}
    \end{subfigure}
    
    \begin{subfigure}{1.6cm}
    \includegraphics[trim=0cm 0cm 0cm 0cm,clip=true,width=1.6cm]{fig/nonGAN/134_skg.jpg}\caption*{\small{Input}}
    \end{subfigure}
    \begin{subfigure}{1.6cm}
    \includegraphics[trim=0cm 0cm 0cm 0cm,clip=true,width=1.6cm]{fig/nonGAN/180_AB.jpg}\vline\caption*{\small{Ref}}
    \end{subfigure}
    \begin{subfigure}{1.6cm}
    \includegraphics[trim=0cm 0cm 0cm 0cm,clip=true,width=1.6cm]{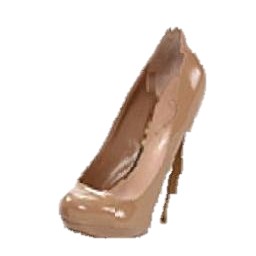}\caption*{\small{SIFT}}
    \end{subfigure}
    \begin{subfigure}{1.6cm}
    \includegraphics[trim=0cm 0cm 0cm 0cm,clip=true,width=1.6cm]{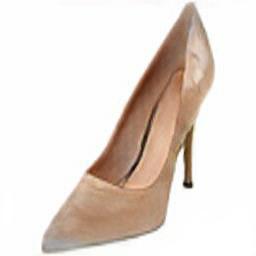}\vline\caption*{\small{$\mathcal{N}$(SIFT)}}
    \end{subfigure}
    \begin{subfigure}{1.6cm}
    \includegraphics[trim=0cm 0cm 0cm 0cm,clip=true,width=1.6cm]{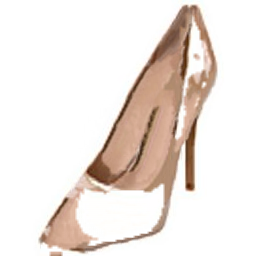}\caption*{\small{DIA}}
    \end{subfigure}
    \begin{subfigure}{1.6cm}
    \includegraphics[trim=0cm 0cm 0cm 0cm,clip=true,width=1.6cm]{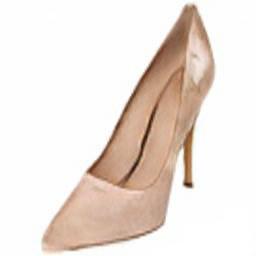}\vline\caption*{\small{$\mathcal{N}$(DIA)}}
    \end{subfigure}
    \begin{subfigure}{1.6cm}
    \includegraphics[trim=0cm 0cm 0cm 0cm,clip=true,width=1.6cm]{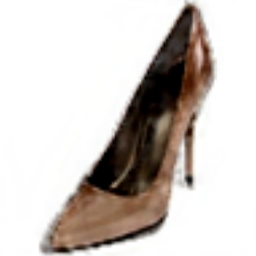}\caption*{\small{PNN}}
    \end{subfigure}
    \begin{subfigure}{1.6cm}
    \includegraphics[trim=0cm 0cm 0cm 0cm,clip=true,width=1.6cm]{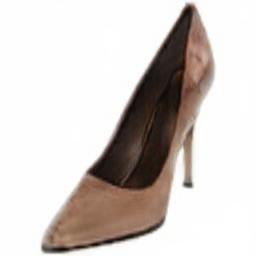}\vline\caption*{\small{$\mathcal{N}$(PNN)}}
    \end{subfigure}
    \begin{subfigure}{1.6cm}
    \includegraphics[trim=0cm 0cm 0cm 0cm,clip=true,width=1.6cm]{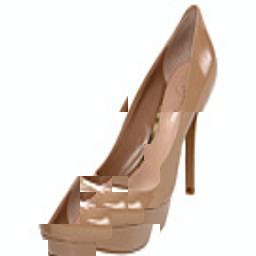}\caption*{\small{SM}}
    \end{subfigure}
    \begin{subfigure}{1.6cm}
    \frame{\includegraphics[trim=0cm 0cm 0cm 0cm,clip=true,width=1.6cm]{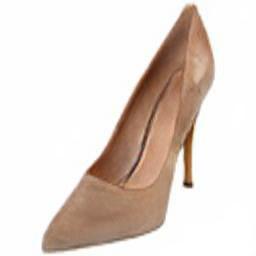}}\caption*{\small{$\mathcal{N}$(SM)}}
    \end{subfigure}
     \caption{\small{The ablation study for the matching methods. We adopt the outputs of SIFT \cite{sift}, DIA \cite{dia} and PNN \cite{pnn} as the matching results and feed them into the translation network $\mathcal{N}$ to demonstrate that $\mathcal{N}$ can generate reasonable results even with other matching methods.}}
     \label{fig:abl_match}
 \end{figure*}

\subsection{Quantitative Evaluation}
\begin{table*}[!ht]
\begin{tabular}{l|cccc|cccc}
\hline
 & \multicolumn{4}{c|}{\textbf{edges $\rightarrow$ shoes}} & \multicolumn{4}{c}{\textbf{edges $\rightarrow$ bags}} \\ \hline
Method & FID & LPIPS & LPIPS (ref) & LPIPS (gt) & FID & LPIPS & LPIPS (ref) & LPIPS (gt) \\ \hline
\tt{pix2pix+noise\cite{pix2pix}} & 75.442 & 0.012 & - & 0.188 & 83.241 & 0.012 & - & 0.253 \\
\tt{BiCycleGAN\cite{bicyclegan}} & 62.521 & 0.108 & - & 0.152 & 70.117 & 0.162 & - & 0.237 \\
\tt{MUNIT\cite{munit}} & 80.857 & 0.110 & - & 0.164 & 88.641 & 0.171 & - & 0.239 \\
\tt{MUNIT+ref} & 80.873 & 0.104 & 0.467 & 0.164 & 89.273 & 0.158 & 0.650 & 0.239 \\
\tt{SEGIN (ours)} & \textbf{54.514} & \textbf{0.145} & \textbf{0.382} & \textbf{0.112} & \textbf{68.211} & \textbf{0.211} & \textbf{0.487} & \textbf{0.158} \\
\tt{real data} & - & 0.301 & - & 0.000 & - & 0.423 & - & 0.000 \\ \hline
\end{tabular}
\caption{\small{LPIPS distance (the higher the better) indicates the output diversities given input and reference. Paired LIPIS scores (the lower the better) including LPIPS (ref) and LPIPS (gt) that shows the similarity between generated images and the reference and reconstruction capability of models, respectively. SEGIN achieves all the best results, which are also highlighted in this table.}}
\label{tbl:eval}
\end{table*}

We perform a quantitative evaluation on the edges2photos dataset for the baseline models and the SEGIN to analyze the generated results in terms of diversity, reconstruction capability, and similarities to the reference.\par
\textbf{Realistic} To evaluate the authentic of our generated results, we use the Fr$\Acute{e}$chet Inception Distance FID score \cite{fid} to measure the distribution similarity between our results and the ground truth dataset, and results are shown in Table.\ref{tbl:eval}. It can be observed that results generated by the SEGIN achieve the best score.

\textbf{Diversity} We use LPIPS metric\cite{lpips} (AlexNet\cite{alexnet} based) to measure the translation diversity. Similar to \cite{bicyclegan}, we compute the average LPIPS distance between pairs of randomly-sampled translation output from the same input. 19 pairs of sampled outputs with different references from 100 inputs are tested in both edge2shoes and edge2handbag dataset. The LPIPS scores are shown in Table \ref{tbl:eval}, and we observe that with various references, the SEGIN achieves the highest score in the experiments. Pix2pix with noise can only produce outputs with a slight variance while BiCycleGAN and MUNIT can achieve similar scores and present comparatively lower diversity compared with the SEGIN.\par
We also found that modal collapse happened in prior models, given reference images have similar styles shown in the $4th$ column (MUNIT 1 and MUNIT 2) in Fig.\ref{fig:comp}. To understand this, we analyze the latent space on the MUNIT. The style code $\Vec{S}$ is extracted by the encoder in the MUNIT and it seems that this specific modal collapse is caused by the collapse of the style latent space since the vector-like can only store limited information. With similar reference, specifically, the black shoe reference in the $4th$ column in Fig.\ref{fig:comp}, their extracted style codes are extremely alike. Since our results preserve the spatial information obtained by semantic-match, the final outputs can be of obvious difference. We compute the LPIPS (ref) score for both MUNIT and the SEGIN with 20 manually selected similar references for 200 input on the edge2shoes dataset and compare the results guided by the same reference at a time. The style distance of MUNIT and the LPIPS (ref) score of both MUNIT and the SEGIN are shown in Table\ref{tbl:eval}, it can be observed that the SEGIN can remain relatively higher diversity even with the similar reference.\par
\textbf{Similarity} To evaluate the similarity between generated images $G(x, r)$ and the reference $r$, we compute the average LPIPS (ref) score for pairs of images between the outputs and the reference, 200 image pairs are tested with 10 references on the edge2photos dataset. In this test, we only compare with MUNIT that supports reference guided style control as well. Mostly, the reference may differ from the outputs in the shape and structure; the SEGIN can still achieve a relatively low score (a lower score indicates smaller distance) of 0.382 compared with MUNIT+ref (0.467) in the edge2shoes dataset, and it also outperforms MUNIT model on the edge2handbags dataset with a lower LPIPS (ref) score.\par

\textbf{Reconstruction Capability} We also evaluate the reconstruction capability for all baseline models and perceive the SEGIN as a quantitative way to judge the visual realism of the results. For Pix2pix and the BiCycleGAN, we compute the LPIPS\cite{lpips} scores between the ground truth $y$ and the generated results, where BiCycleGAN generates images with the encoded style data. As for MUNIT and the SEGIN, we use the ground truth as a reference to guide the models to generate images resembling the ground truth to the full extent, and then we evaluate the LPIPS between paired generated images and the ground truth. In Table \ref{tbl:eval}, our model SEGIN outperforms these baselines in terms of the reconstruction quality.\par

\subsection{Ablation Study}
\begin{figure*}[ht]
     \centering
     \begin{subfigure}{1.6cm}
    \includegraphics[trim=0cm 0cm 0cm 0cm,clip=true,width=1.6cm]{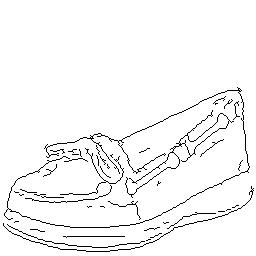}
    \end{subfigure}
    \begin{subfigure}{1.6cm}
    \includegraphics[trim=0cm 0cm 0cm 0cm,clip=true,width=1.6cm]{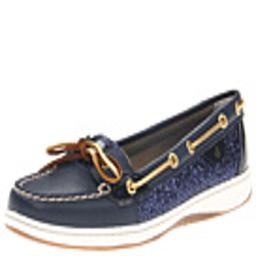}
    \end{subfigure}
    \begin{subfigure}{1.6cm}
    \includegraphics[trim=0cm 0cm 0cm 0cm,clip=true,width=1.6cm]{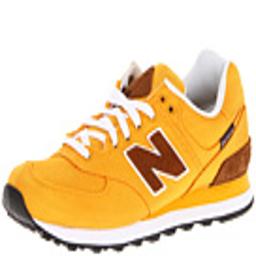}
    \end{subfigure}
    \begin{subfigure}{1.6cm}
    \frame{\includegraphics[trim=0cm 0cm 0cm 0cm,clip=true,width=1.6cm]{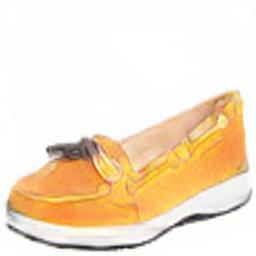}}
    \end{subfigure}
    \begin{subfigure}{1.6cm}
    \includegraphics[trim=0cm 0cm 0cm 0cm,clip=true,width=1.6cm]{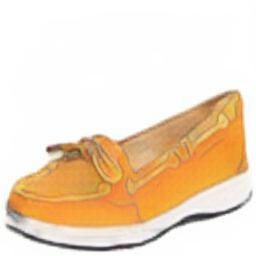}
    \end{subfigure}
    \begin{subfigure}{1.6cm}
    \includegraphics[trim=0cm 0cm 0cm 0cm,clip=true,width=1.6cm]{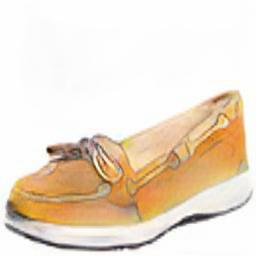}
    \end{subfigure}
    \begin{subfigure}{1.6cm}
    \includegraphics[trim=0cm 0cm 0cm 0cm,clip=true,width=1.6cm]{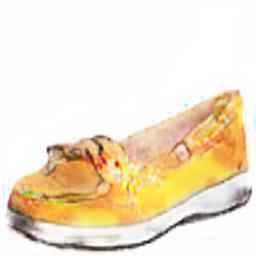}
    \end{subfigure}
    \begin{subfigure}{1.6cm}
    \includegraphics[trim=0cm 0cm 0cm 0cm,clip=true,width=1.6cm]{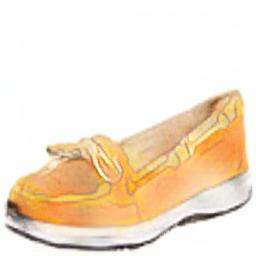}
    \end{subfigure}
    \begin{subfigure}{1.6cm}
    \includegraphics[trim=0cm 0cm 0cm 0cm,clip=true,width=1.6cm]{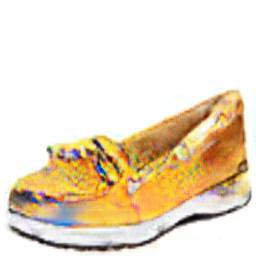}
    \end{subfigure}
    \begin{subfigure}{1.6cm}
    \includegraphics[trim=0cm 0cm 0cm 0cm,clip=true,width=1.6cm]{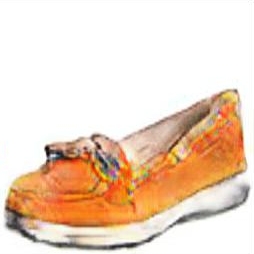}
    \end{subfigure}
    
    \begin{subfigure}{1.6cm}
    \includegraphics[trim=0cm 0cm 0cm 0cm,clip=true,width=1.6cm]{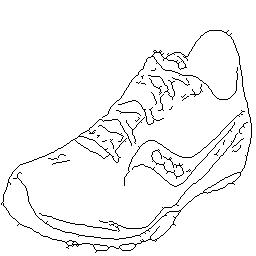}
    \end{subfigure}
    \begin{subfigure}{1.6cm}
    \includegraphics[trim=0cm 0cm 0cm 0cm,clip=true,width=1.6cm]{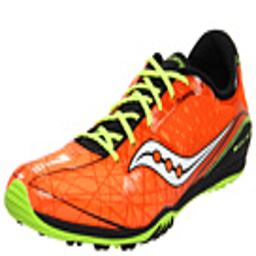}
    \end{subfigure}
    \begin{subfigure}{1.6cm}
    \includegraphics[trim=0cm 0cm 0cm 0cm,clip=true,width=1.6cm]{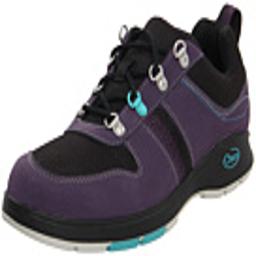}
    \end{subfigure}
    \begin{subfigure}{1.6cm}
    \frame{\includegraphics[trim=0cm 0cm 0cm 0cm,clip=true,width=1.6cm]{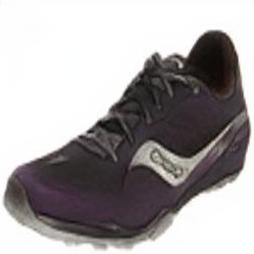}}
    \end{subfigure}
    \begin{subfigure}{1.6cm}
    \includegraphics[trim=0cm 0cm 0cm 0cm,clip=true,width=1.6cm]{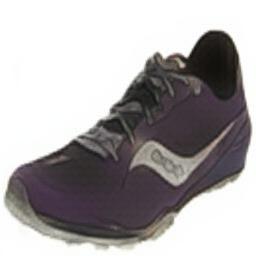}
    \end{subfigure}
    \begin{subfigure}{1.6cm}
    \includegraphics[trim=0cm 0cm 0cm 0cm,clip=true,width=1.6cm]{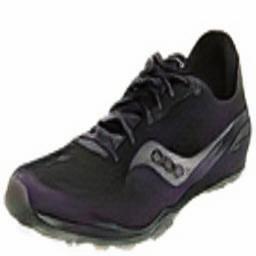}
    \end{subfigure}
    \begin{subfigure}{1.6cm}
    \includegraphics[trim=0cm 0cm 0cm 0cm,clip=true,width=1.6cm]{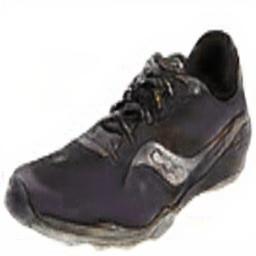}
    \end{subfigure}
    \begin{subfigure}{1.6cm}
    \includegraphics[trim=0cm 0cm 0cm 0cm,clip=true,width=1.6cm]{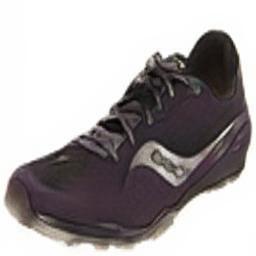}
    \end{subfigure}
    \begin{subfigure}{1.6cm}
    \includegraphics[trim=0cm 0cm 0cm 0cm,clip=true,width=1.6cm]{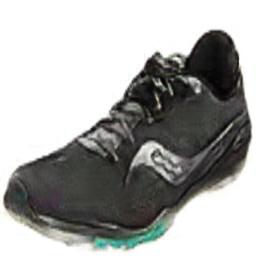}
    \end{subfigure}
    \begin{subfigure}{1.6cm}
    \includegraphics[trim=0cm 0cm 0cm 0cm,clip=true,width=1.6cm]{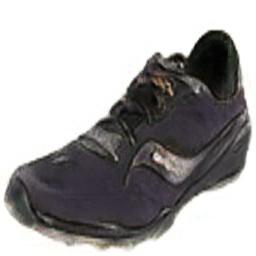}
    \end{subfigure}
    
   \begin{subfigure}{1.6cm}
    \includegraphics[trim=0cm 0cm 0cm 0cm,clip=true,width=1.6cm]{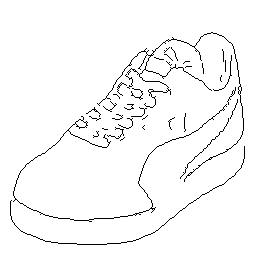}
    \end{subfigure}
    \begin{subfigure}{1.6cm}
    \includegraphics[trim=0cm 0cm 0cm 0cm,clip=true,width=1.6cm]{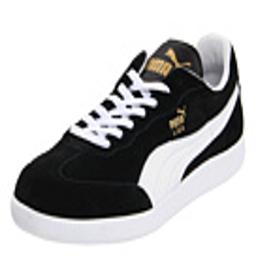}
    \end{subfigure}
    \begin{subfigure}{1.6cm}
    \includegraphics[trim=0cm 0cm 0cm 0cm,clip=true,width=1.6cm]{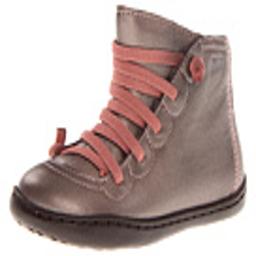}
    \end{subfigure}
    \begin{subfigure}{1.6cm}
    \frame{\includegraphics[trim=0cm 0cm 0cm 0cm,clip=true,width=1.6cm]{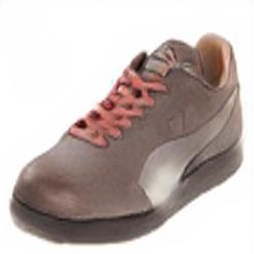}}
    \end{subfigure}
    \begin{subfigure}{1.6cm}
    \includegraphics[trim=0cm 0cm 0cm 0cm,clip=true,width=1.6cm]{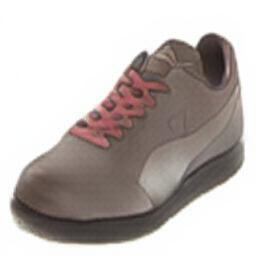}
    \end{subfigure}
    \begin{subfigure}{1.6cm}
    \includegraphics[trim=0cm 0cm 0cm 0cm,clip=true,width=1.6cm]{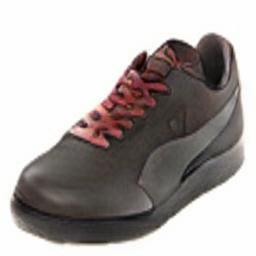}
    \end{subfigure}
    \begin{subfigure}{1.6cm}
    \includegraphics[trim=0cm 0cm 0cm 0cm,clip=true,width=1.6cm]{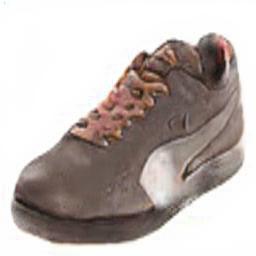}
    \end{subfigure}
    \begin{subfigure}{1.6cm}
    \includegraphics[trim=0cm 0cm 0cm 0cm,clip=true,width=1.6cm]{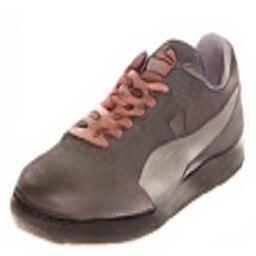}
    \end{subfigure}
    \begin{subfigure}{1.6cm}
    \includegraphics[trim=0cm 0cm 0cm 0cm,clip=true,width=1.6cm]{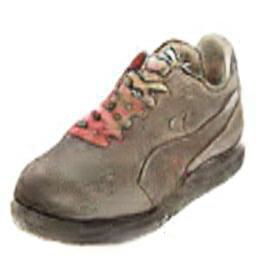}
    \end{subfigure}
    \begin{subfigure}{1.6cm}
    \includegraphics[trim=0cm 0cm 0cm 0cm,clip=true,width=1.6cm]{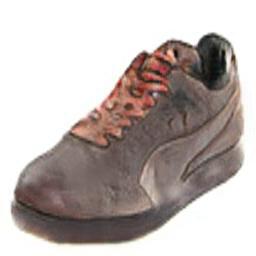}
    \end{subfigure}
    
    \begin{subfigure}{1.6cm}
    \includegraphics[trim=0cm 0cm 0cm 0cm,clip=true,width=1.6cm]{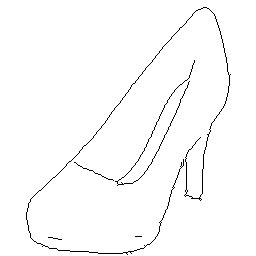}\caption*{\small{Input}}
    \end{subfigure}
    \begin{subfigure}{1.6cm}
    \includegraphics[trim=0cm 0cm 0cm 0cm,clip=true,width=1.6cm]{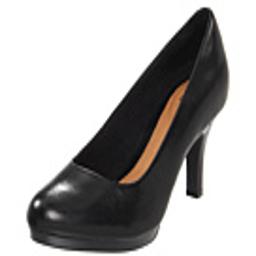}\caption*{\small{GT}}
    \end{subfigure}
    \begin{subfigure}{1.6cm}
    \includegraphics[trim=0cm 0cm 0cm 0cm,clip=true,width=1.6cm]{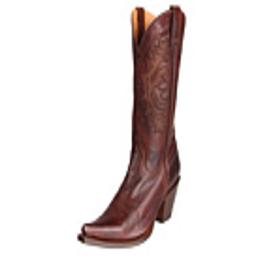}\caption*{\small{Ref}}
    \end{subfigure}
    \begin{subfigure}{1.6cm}
    \frame{\includegraphics[trim=0cm 0cm 0cm 0cm,clip=true,width=1.6cm]{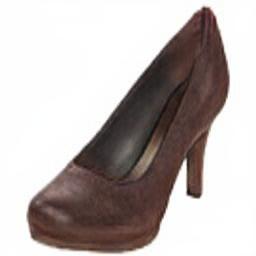}}\caption*{\small{Output}}
    \end{subfigure}
    \begin{subfigure}{1.6cm}
    \includegraphics[trim=0cm 0cm 0cm 0cm,clip=true,width=1.6cm]{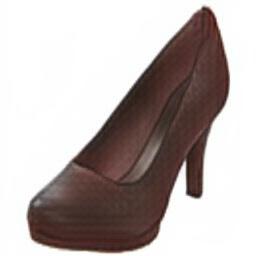}\caption*{\small{w/o $L_{GAN}$}}
    \end{subfigure}
    \begin{subfigure}{1.6cm}
    \includegraphics[trim=0cm 0cm 0cm 0cm,clip=true,width=1.6cm]{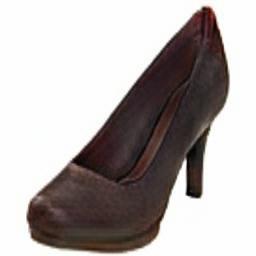}\caption*{\small{w/o $L_{Recon}$}}
    \end{subfigure}
    \begin{subfigure}{1.6cm}
    \includegraphics[trim=0cm 0cm 0cm 0cm,clip=true,width=1.6cm]{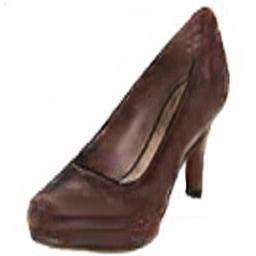}\caption*{\small{w/o $L_{F}$}}
    \end{subfigure}
    \begin{subfigure}{1.6cm}
    \includegraphics[trim=0cm 0cm 0cm 0cm,clip=true,width=1.6cm]{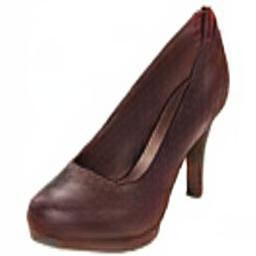}\caption*{\small{w/o $L_{tv}$}}
    \end{subfigure}
    \begin{subfigure}{1.6cm}
    \includegraphics[trim=0cm 0cm 0cm 0cm,clip=true,width=1.6cm]{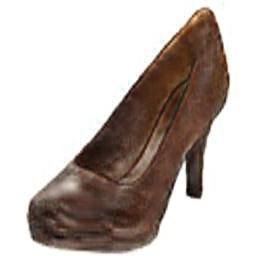}\caption*{\small{w/o NL}}
    \end{subfigure}
    \begin{subfigure}{1.6cm}
    \includegraphics[trim=0cm 0cm 0cm 0cm,clip=true,width=1.6cm]{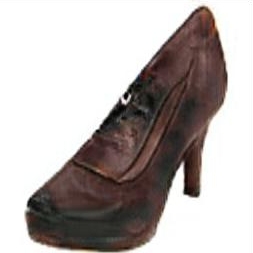}\caption*{\small{w/ MT}}
    \end{subfigure}
     \caption{\small{The ablation study of losses}}
     \label{fig:abl}
 \end{figure*}

We present an ablation study for the losses used in our model, shown in Fig.\ref{fig:abl}. It can be observed that without the GAN loss, the results are flat and lack of realism though it can maintain most of the features. Output without $L_{Recon}$ and $L_{F}$ is shown in the $6^{th}$ and the $7^{th}$ column, respectively; apparent color differences between the generated images and reference images can be noticed when the reconstruction loss is not employed, while sever artifacts can be observed without $L_{F}$. Compared with those major loss functions, the TV loss has a relatively slighter impact on the results, but it does help to smooth the images and reduce incoherent variation within one image. The color in the yellow shoes (the first row) changed sharply when the TV loss is not used, and the result seems like being mottled with white and yellow patches while the result with TV loss is pretty natural and of color consistency. Quantitative results are included in the ablation study as well. We use the Frechet Inception Distance (FID) \cite{fid} to measure the distribution of generated outputs and the real images. Meanwhile, we calculate the LPIPS pair score mentioned previously to assess the reconstruction ability of the model without applying certain loss functions, results of which are shown in Table. \ref{tbl:abl}.\par
 
\begin{table}[!ht]
\begin{tabular}{l|cc|cc}
\hline
 & \multicolumn{2}{c|}{\textbf{edges $\rightarrow$ shoes}} & \multicolumn{2}{c}{\textbf{edges $\rightarrow$ bags}} \\ \hline
\multicolumn{1}{c|}{Method} & \multicolumn{1}{c}{FID} & \multicolumn{1}{c|}{LPIPS (gt)} & \multicolumn{1}{c}{FID} & \multicolumn{1}{c}{LPIPS (gt)} \\ \hline
w/o GAN & 55.618 & 0.112 & 69.981 & 0.158 \\ \hline
w/o $L_{recon}$ & 59.162 & 0.121 & 78.301 & 0.174 \\ \hline
w/o $L_{F}$ & 76.554 & 0.147 & 96.454 & 0.211 \\ \hline
w/o $L_{tv}$ & 58.181 & \textbf{0.111} & 72.538 & \textbf{0.155} \\ \hline
w/o NL & 72.387 & 0.132 & 94.113 & 0.172 \\ \hline
w/o MT & 77.211 & 0.144 & 98.240 & 0.181 \\ \hline
\textbf{SEGIN} & \textbf{54.514} & 0.112 & \textbf{68.211} & 0.158 \\ \hline
\end{tabular}
\caption{\small{FID and LPIPS Pair score of results in the ablation study. The SEGIN achieves the best FID scores in both datasets as well as relatively lower LPIPS (gt) score compared with most of other methods. Results without TV loss $L_{tv}$ obtained the lowest LPIPS (gt) score since the smoothness may cause differences during reconstruction. However, with the TV loss, the overall image looks more authentic and achieves much better FID score.}}
\label{tbl:abl}
\end{table}
We also conduct ablation experiments to demonstrate that the non-local block and the multi-task framework have an essential contribution to improve the output in terms of color propagation as well as the generation of the complete object boundary. We notice that the color cannot be propagated properly in the results generated without the non-local layers. While there is no multi-task framework that can generate the attention mask, the output disposes of artifacts around the edge of objects. We also found those artifacts in prior works \cite{pix2pix,munit}, and it can be eliminated by generating the attention mask of auxiliary information. We show some ablation study results in Fig.\ref{fig:abl}\par

It is also possible to use different match methods to replace the semantic match module. Traditional methods to locate the correspondence, such as SIFT Flow \cite{sift} and results generated by DIA and PNN can also be deemed as the corresponding matching produced in different ways. We perform an ablation study to show that our network has the robustness for reasonable outputs even with different matching methods, while current semantic match generates the best results. As shown in Fig.\ref{fig:abl_match}, we feed the results of SIFT \cite{sift}, DIA \cite{dia} and PNN \cite{pnn} into our translation network severally as auxiliary images. It can be observed that the output images are still of high quality even some of those auxiliary images are invalid or blurry.\par

\section{Conclusion}
In this paper, we propose a novel I2I translation method that the generated outputs can not only learn a many-to-many mapping between two visual domains but also be guided by reference images semantically. We present a semantic match approach to find corresponding matched patches between two visual contents and model the generator with a multi-task framework as well as non-local layers, which is trained in a self-supervised manner. Both qualitative and quantitative evaluations demonstrate that our method produces realistic and diverse results with higher semantically similarity to the reference compared with the state-of-the-art prior works. However, there are also limitations in this work: (1) both input and the reference are required to contain semantic information; (2) when the semantic relation between the input and the reference is weak, the translation will be degraded with only global guidance, and the failure case can be found in the Appendix.  

\bibliography{output}
\includegraphics[width=0.2\textwidth]{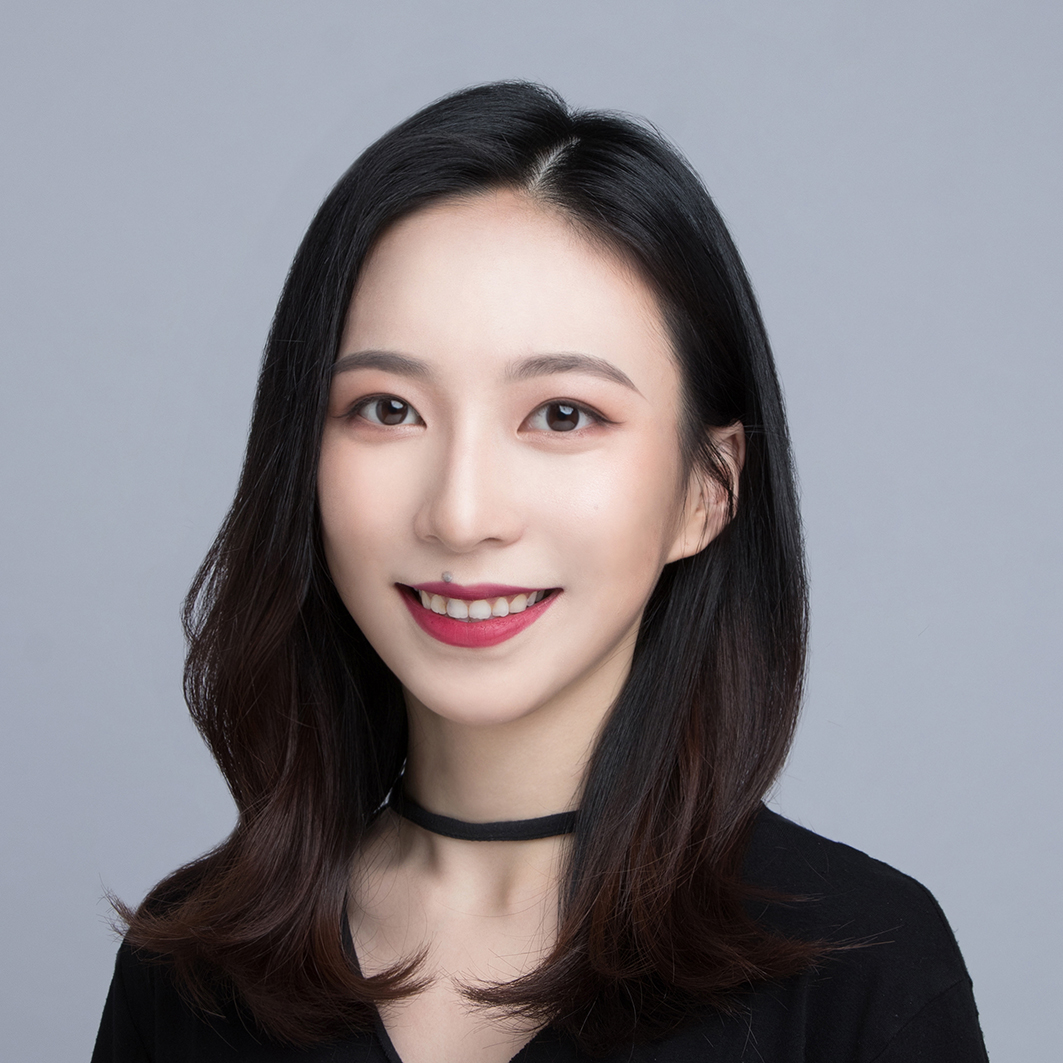}\textbf{Jialu Huang} is a PhD student in the Department of Computer Science, City University of Hong Kong (CityU) since Sep 2018. Prior to that, she was a research developer in ASML. She received the B.Eng. degree from Sun Yat-Sen University and MSc. degrees from King's College London. Her primary research interests fall in the fields of Image synthesis, Computer Graphics, Evolutionary Algorithm.\par
\includegraphics[width=0.2\textwidth]{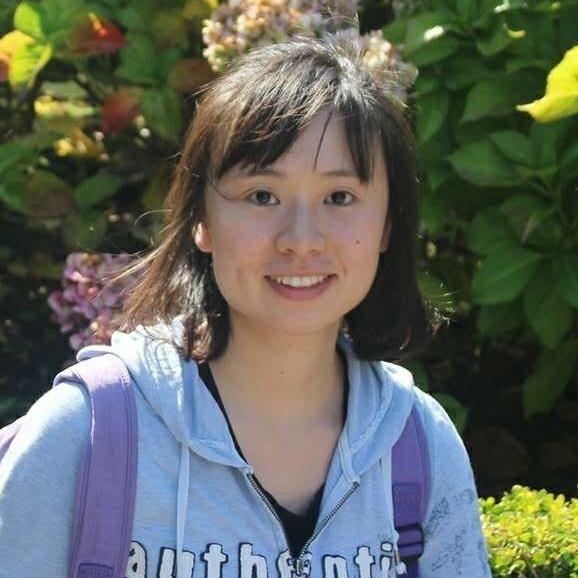}\textbf{Dr. Liao Jing} is an Assistant Professor with the Department of Computer Science, City University of Hong Kong (CityU) since Sep 2018. Prior to that, she was a Researcher at Visual Computing Group, Microsoft Research Asia from 2015 to 2018. She received the B.Eng. degree from HuaZhong University of Science and Technology and dual Ph.D. degrees from Zhejiang University and Hong Kong UST. Her primary research interests fall in the fields of Computer Graphics, Computer Vision, Image/Video Processing, Digital Art and Computational Photography.\par
\includegraphics[width=0.2\textwidth]{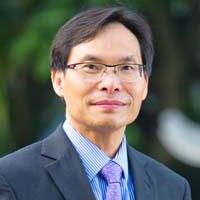}\textbf{Prof. Kwong Tak Wu Sam} joined City University as a lecturer in the Department of Electronic Engineering in 1989. Before joining City University, he worked for Control Data Canada and Bell Northern Research as diagnostic engineer and member of Scientific Staff, respectively. At present, he is the associate editor of the IEEE transactions on Industrial Informatics and IEEE Transactions on Industrial Electronics, Journal of Information Sciences. He is also the Admissions Officer for the graduate programme in the Department. His research interests are evolutionary algorithms, pattern recognition, digital watermarking, video coding and network intrusion systems.\par
\end{document}